\documentclass[sigconf]{acmart}

\usepackage{amsmath,amsfonts,bm}

\def\eqref#1{equation~\ref{#1}}

\def\1{\bm{1}}

\DeclareMathAlphabet{\mathsfit}{\encodingdefault}{\sfdefault}{m}{sl}
\SetMathAlphabet{\mathsfit}{bold}{\encodingdefault}{\sfdefault}{bx}{n}

\usepackage{url}
\usepackage{graphicx}
\usepackage{adjustbox}
\usepackage{pdflscape}
\usepackage{tikz}
\usetikzlibrary{arrows.meta, positioning}
\usepackage{listings}
\definecolor{navyblue}{RGB}{0,0,128}
\usepackage{amsmath}
\usepackage{enumitem}

\usepackage{amssymb}
\usepackage{algorithm}
\usepackage{algpseudocode}
\usepackage{rotating}

\usepackage{xcolor}
\usepackage{float}
\usepackage{booktabs}       %
\usepackage{microtype}
\usepackage{multirow}
\UseRawInputEncoding
\usepackage{tcolorbox}

\definecolor{darkblue}{rgb}{0, 0, 0.5}
\hypersetup{colorlinks=true, citecolor=darkblue, linkcolor=darkblue, urlcolor=darkblue, pdfcontactemail={}, pdfcontacturl={}, pdfcontactaddress={}}

\lstdefinelanguage{prompt}{
  morekeywords={Role,Inputs,Procedure,Output,Scoring,Helpfulness,Truthfulness,Completeness},
  sensitive=true
}

\lstdefinestyle{promptstyle}{
  language=prompt,
  basicstyle=\ttfamily\footnotesize,
  columns=fullflexible,
  breaklines=true,
  showstringspaces=false,
  frame=single,
  rulecolor=\color{black!25},
  backgroundcolor=\color{gray!6},
  xleftmargin=1.1em,
  framexleftmargin=1.1em,
  aboveskip=0.6\baselineskip,
  belowskip=0.6\baselineskip,
  numbers=left,
  numberstyle=\scriptsize\color{black!40},
  keywordstyle=\bfseries\color{blue!60!black},
  emphstyle=\bfseries\color{purple!60!black},
}

\lstdefinestyle{dspystyle}{
  language=Python,
  basicstyle=\ttfamily\footnotesize,
  columns=fullflexible,
  breaklines=true,
  showstringspaces=false,
  frame=single,
  rulecolor=\color{black!25},
  backgroundcolor=\color{gray!6},
  xleftmargin=1.1em,
  framexleftmargin=1.1em,
  numbers=left,
  numberstyle=\scriptsize\color{black!40},
  keywordstyle=\bfseries\color{blue!60!black},
  commentstyle=\color{green!40!black},
  stringstyle=\color{black!75},
}

\newcommand{\gpt}{\texttt{GPT-5}}
\newcommand{\gpto}{\texttt{gpt-oss-20b}}
\newcommand{\gptf}{\texttt{GPT-4.1}}

\newcommand{\pytorch}{\texttt{PyTorch}}
\newcommand{\neobert}{\texttt{NeoBERT}}

\newcommand{\reasm}{\texttt{RM-R1-Qwen-14B}}
\newcommand{\mentat}{\texttt{MENTAT}}
\newcommand{\gepa}{\texttt{GEPA}}
\newcommand{\zooter}{\texttt{Zooter}}
\newcommand{\routellm}{\texttt{RouteLLM}}
\newcommand{\llmblender}{\texttt{LLM-Blender}}

\setcopyright{acmlicensed}

\copyrightyear{2026}
\acmYear{2026}
\setcopyright{cc}
\setcctype{by}
\acmConference[CAIS '26]{ACM Conference on AI and Agentic Systems}{May 26--29, 2026}{San Jose, CA, USA}
\acmBooktitle{ACM Conference on AI and Agentic Systems (CAIS '26), May 26--29, 2026, San Jose, CA, USA}
\acmDOI{10.1145/3786335.3813139}
\acmISBN{979-8-4007-2415-2/2026/05}

\begin{document}
\title{Reasoning-Intensive Regression}

\author{Diane Tchuindjo}
\affiliation{
  \institution{Massachusetts Institute of Technology}
  \city{Cambridge}
  \state{MA}
  \country{USA}
}
\email{dianetc@mit.edu}

\author{Omar Khattab}
\affiliation{
  \institution{Massachusetts Institute of Technology}
  \city{Cambridge}
  \state{MA}
  \country{USA}
}
\email{okhattab@mit.edu}

\begin{abstract}

AI researchers and practitioners increasingly apply large language models (LLMs) to what we call reasoning-intensive regression (RiR), i.e., deducing subtle numerical scores from text. Unlike standard language regression tasks such as sentiment or similarity analysis, RiR often appears instead in ad-hoc applications such as rubric-based scoring, modeling dense rewards in complex environments, or domain-specific retrieval, where much deeper analysis of context is required while only limited task-specific training data and computation are available. We cast four realistic problems as RiR tasks to establish an initial benchmark, and use that to test our hypothesis that prompting frozen LLMs and fine-tuning Transformer encoders via gradient descent will both often struggle in RiR. We then propose \mentat{}, a simple and lightweight method that combines batch-reflective prompt optimization with neural ensemble learning.  \mentat{} achieves up to $65\%$ improvement over both baselines, though substantial room remains for future advances.\footnote{Data: \url{https://huggingface.co/datasets/dianetc/rir-paper-data}}

\end{abstract}

\maketitle

\section{Introduction}
\label{introduction}

\begin{figure}[t]
    \centering
   \includegraphics[width=237pt]{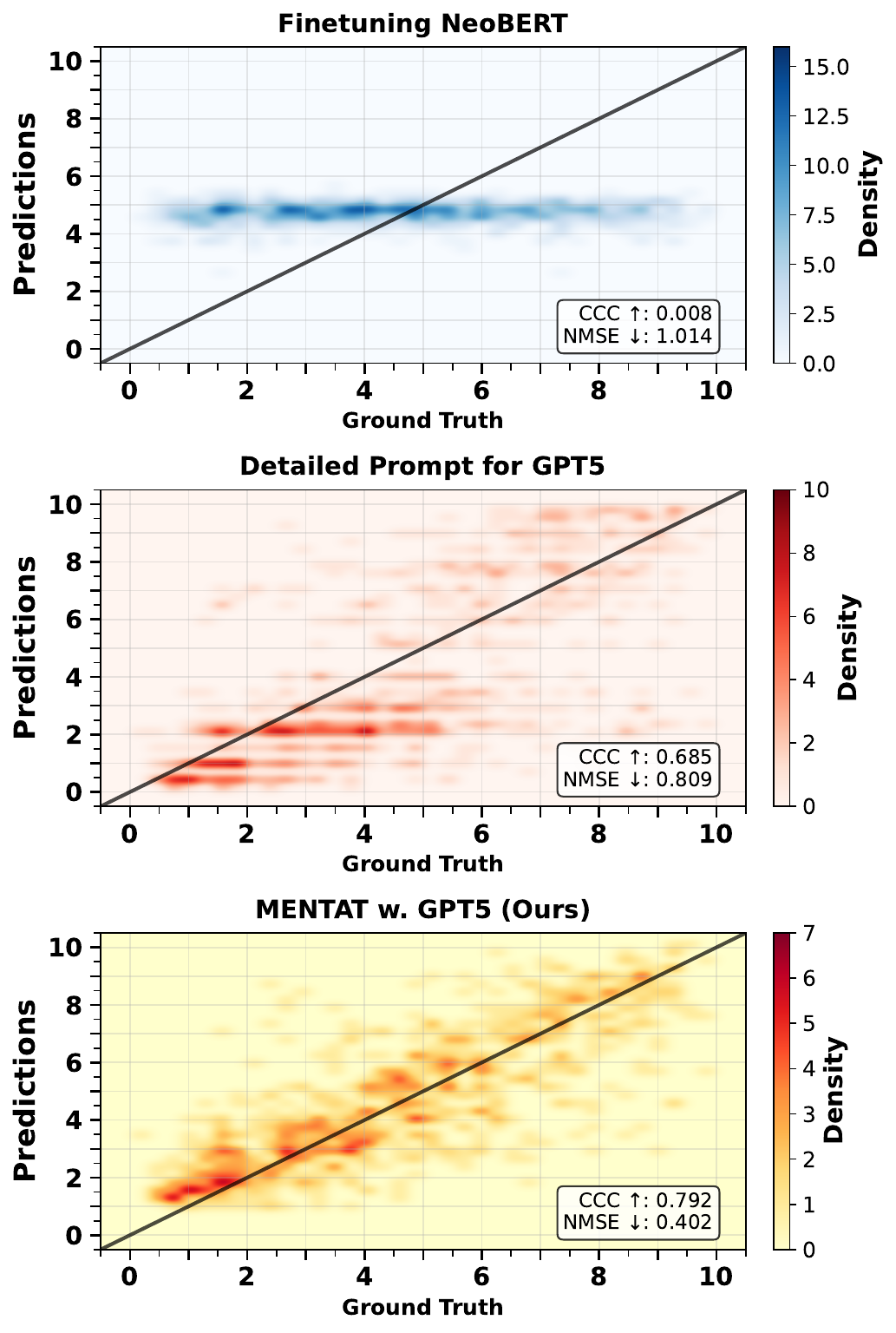}
\vspace{-2mm}
\caption{On regression for detecting the first math error in a reasoning trace, fine-tuning a NeoBERT model collapses and human-crafted prompting exhibits coarse, imprecise predictions. \mentat's performance illustrates how RiR problems benefit from combining deep reasoning capabilities with precise numerical predictions.
}
\vspace{-5mm}
\label{fig:mentat_v_others}
\end{figure}

Despite fast progress in adapting large language models (LLMs) for downstream problems, lightweight methods for teaching LLMs to do even standard \textit{natural-language regression} tasks remain surprisingly elusive~\citep{regression-aware, anonymous2024better,embeddingsregression,omnipred,song2025decodingbasedregression}. These tasks, like sentiment analysis, semantic similarity, and document ranking, involve predicting a score \( y \in \mathbb{R} \) from a natural-language string. On these problems, applying straightforward supervised learning to pretrained Transformer encoders such as BERT~\citep{bert} has been shown to perform competitively with much larger decoder-only LLMs~\citep{anonymous2024better}, even with sophisticated fine-tuning methods. 

\begin{figure*}[t]
    \centering
   \includegraphics[height=0.4\textheight]{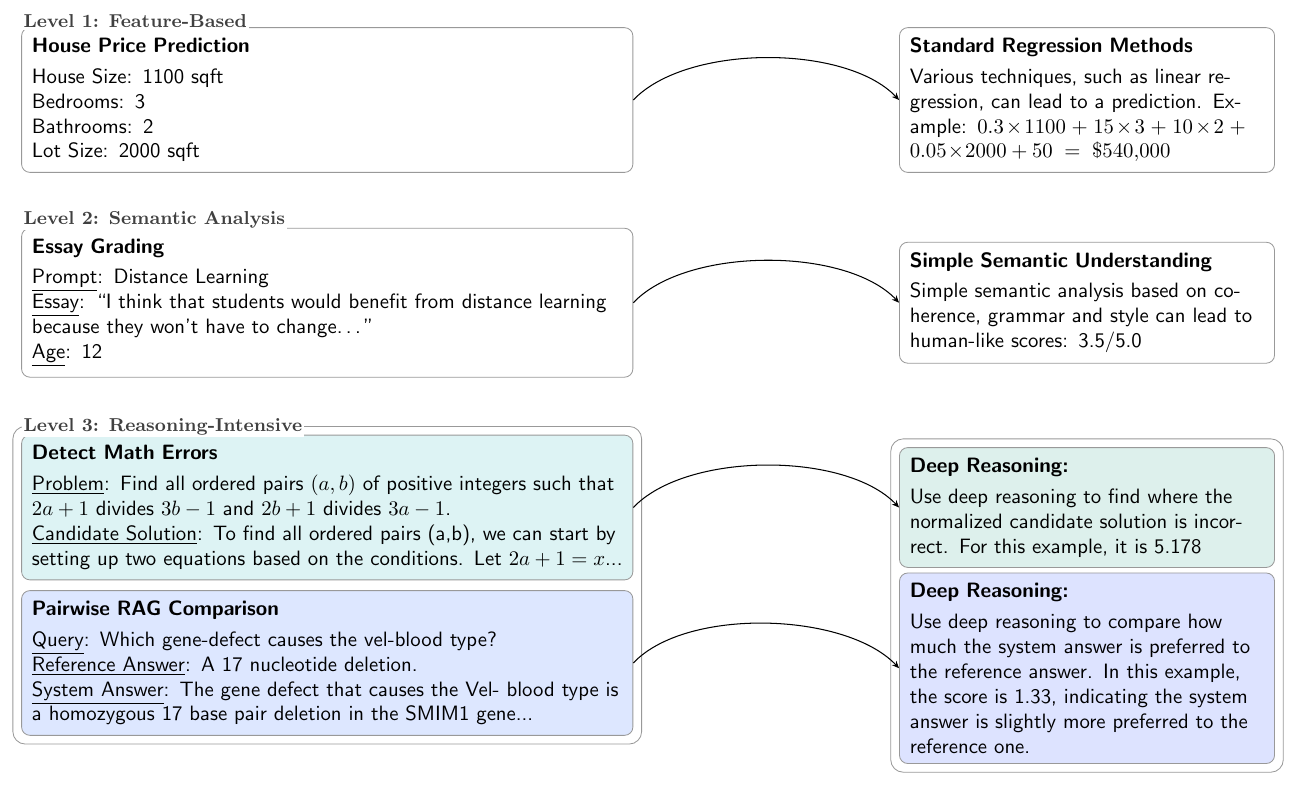} \\
\caption{Inspired by \citet{reasoning-intensive-retrieval}'s analysis of retrieval tasks, we break down text-based regression problems into three, informal complexity levels. Level 1 tasks use simple feature-based inputs (for example, the number of bedrooms and bathrooms when predicting home prices). Text-to-text regression achieves strong Level 1 performance with rich datasets~\citep{text-to-text-regression}. Level 2 tasks require moderate semantic understanding (sentiment analysis, reward modeling) but are easy for supervised-learning over a pretrained Transformer. Level 3, the focus of this work, represents Reasoning-Intensive Regression (RiR), which requires deep sequential reasoning.}
\label{fig:regression-levels}
\end{figure*}

We investigate what we call \textit{Reasoning-Intensive Regression} (RiR), a fuzzy but growing subset of natural-language regression in which processing the text in each instance \textit{demands sequential deduction or deep analysis, rather than shallow identification of features}. Unlike simpler regression tasks, RiR problems call for explicit step-by-step problem decomposition or \textit{reasoning}, where the system produces intermediate sequences of steps like tokens $\langle r_1, ..., r_t \rangle \in \Sigma^*$ before committing to a prediction~\citep{expressivecot}. See Figure~\ref{fig:regression-levels} for a breakdown of regression problems into three levels of complexity: feature-based, semantic analysis, and reasoning-intensive, inspired by \citet{reasoning-intensive-retrieval}'s analysis of retrieval tasks.

These types of applications are emerging rapidly in both research and practice, e.g., to produce scores for ad-hoc applications that process customer calls, student essays, rubric-based LLM generation, or instruction-based query--document relevance~\citep{assuranceiq2024scoring,es2024ragas,reasoning-intensive-retrieval,thakur2025support}. In parallel, the same scoring paradigm is being scaled in recent efforts toward general-purpose chain-of-thought reward models~\citep{team2025kimi,critiqueoutloud}, but these typically assume orders-of-magnitude more labels and compute (e.g., hundreds of thousands of labels in K2) than the lightweight application-specific regimes that are far more common in the long tail.

We establish an initial benchmark for RiR by casting four realistic tasks as regression problems that demand \textit{varying} levels of reasoning: predicting the proportion of a long mathematical deduction up to the first erroneous statement, determining the extent to which an LLM can follow highly composite instructions, predicting the degree to which the response of one Retrieval-Augmented Generation (RAG) system is better than another, and grading student essays on supplied topics. We then identify two practical constraints of downstream RiR applications. Such applications typically offer only very small training sets and are limited to lightweight computations (LLM inference, lightweight prompt optimization, and fine-tuning of medium-sized networks such as small Transformers), precluding approaches like large-scale reinforcement learning for LLMs~\citep{deepseek,team2025kimi}.

We ask: \textit{Are there effective methods that are data- and compute-efficient for tackling ad-hoc reasoning-intensive regression problems?} We hypothesize that what makes RiR problems especially challenging is that they combine the reasoning need for \textit{deep analysis of each individual task instance} with the regression challenge of \textit{learning to produce precise, calibrated, and well-ranked scores} from very little data. As illustrated in Figure~\ref{fig:mentat_v_others}, standard prompt engineering techniques struggle with the high precision needed for learning to approximate a statistical distribution, while approaches that bypass LLM-based reasoning, e.g., training small Transformer encoders, often fail to truly learn RiR problems and instead seek to ``hack'' the regression loss function by finding degenerate approximations (e.g., collapsing to a small range of scores).

\begin{figure*}[t]
    \centering
    \includegraphics[width=0.24\textwidth]{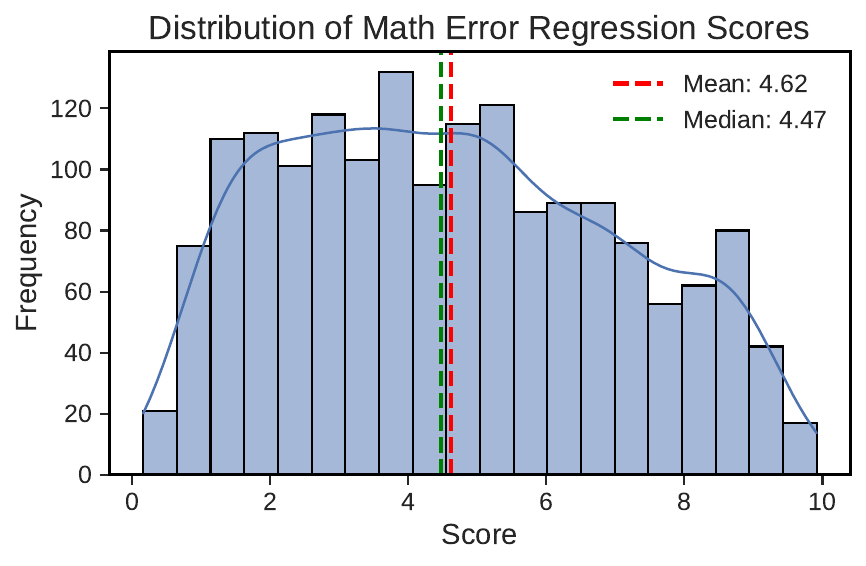}
    \hfill
    \includegraphics[width=0.24\textwidth]{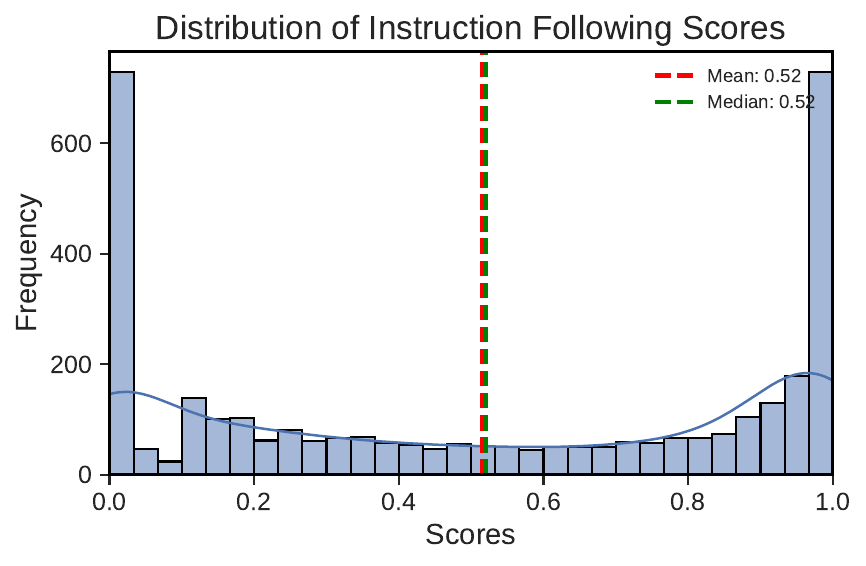}
    \hfill
    \includegraphics[width=0.24\textwidth]{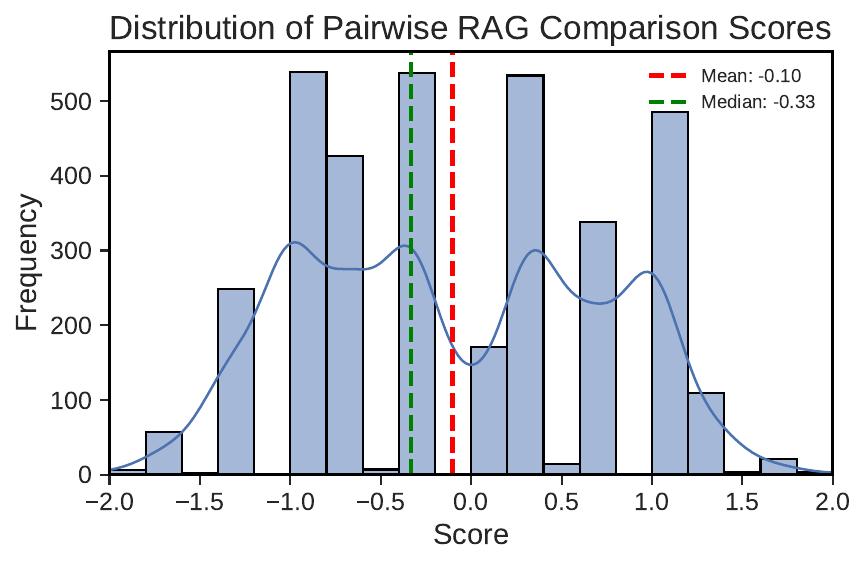}
    \hfill
    \includegraphics[width=0.24\textwidth]{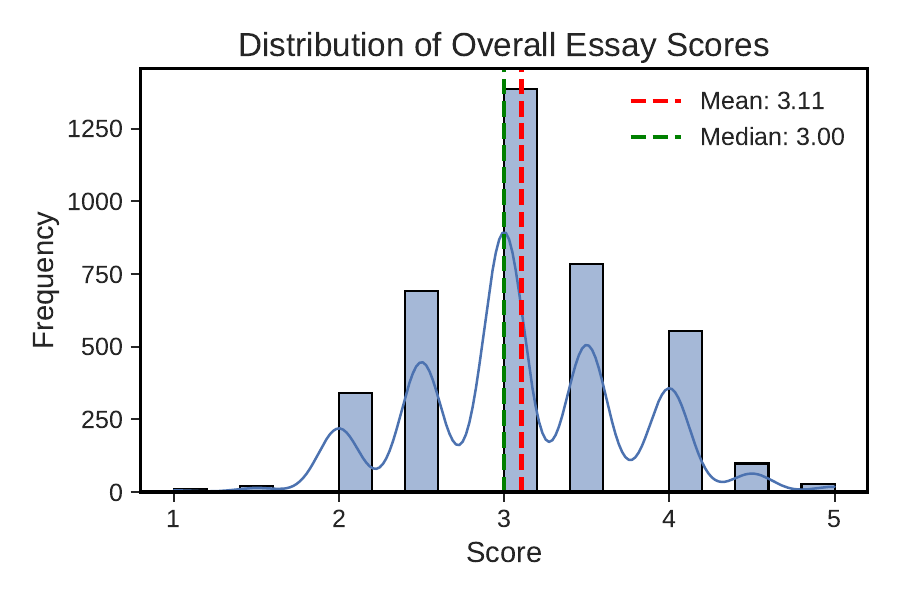}
    \caption{Ground-truth score distributions for mathematical error detection (the spread capturing the tendency for solutions to fail towards the center), instruction following (capturing the tendency to favor the tails), pairwise RAG comparison (narrow distribution around averaged judgments), and essay grading (tight clustering characteristic of qualitative assessments).}
    \label{fig:full-distributions}
\end{figure*}

We propose \textbf{M}istake-Aware prompt \textbf{E}volver with \textbf{N}eural \textbf{T}raining \textbf{A}nd \textbf{T}esting (\mentat{}), a simple and lightweight method that combines iterative prompt optimization with neural regression. Rather than relying on LLMs to produce precise numerical predictions directly, \mentat{} uses an iterative error-driven prompt evolution process. Starting with even just a very basic prompt, the LLM analyzes its own prediction errors \textit{in large batches}, identifies patterns of its poor performance, and then refines the prompt based on that. After few iterations, \mentat{} trains a simple aggregation MLP to reduce multiple rollouts from the LLM-discovered prompt into a final prediction. \mentat{} delivers consistent improvements in quality, but nonetheless leaves large headroom on many of the RiR settings we define. Our contributions are threefold:

\begin{enumerate}[topsep=5pt,itemsep=2pt,parsep=0pt,partopsep=0pt]
    \item {\bf Problem Formulation.} We formalize Reasoning-Intensive Regression (RiR) as a distinct subclass of natural-language regression, distinguishing it from feature-based and semantic-analysis tasks by its requirement for explicit multi-step reasoning before score prediction.
    \item {\bf Benchmark.} We establish an initial RiR benchmark comprising four tasks of varying reasoning intensity: mathematical error detection, instruction following, pairwise RAG comparison, and essay grading. We advocate for Concordance Correlation Coefficient (CCC) as a more appropriate metric than NMSE for RiR evaluation, as it captures both ranking quality and calibration.
    \item {\bf Method and Analysis.} We propose MENTAT, a lightweight method combining batch-reflective prompt optimization with neural ensemble learning. Through systematic evaluation, we demonstrate that neither prompting frozen LLMs nor fine-tuning Transformer encoders alone suffices for RiR, while MENTAT achieves up to 65\% improvement over baselines, though substantial headroom remains.
\end{enumerate}

The remainder of the study is as follows: Section 2 describes how we translate four problems into RiR tasks and Section 3 introduces \mentat{}. Section 4 presents our evaluation methodology, including the details of our baselines, and the results. The paper concludes with Section 5, which synthesize our findings and discuss implications for future research. An extended discussion of related work is given in the Appendix~\ref{related-works}.

\section{Benchmarking RiR}
We collect four tasks for Reasoning-Intensive Regression, chosen to span a range of reasoning intensities (from pairwise preference judgment to mathematical error localization) and label distributions (bimodal, narrow, tight clustering).  Refer to Figure \ref{fig:full-distributions} for dataset distributions. We view these as representative rather than exhaustive.

\begin{itemize}
    \item {\bf Mathematical Error Detection} requires precise logical reasoning and stepwise analysis, while also stressing the fact that LLMs are known to struggle with precisely estimating simple properties like text length.
    \item {\bf Instruction Following} evaluates how well a response satisfies a set of fine-grained requirements, and expects models to produce calibrated scalar judgments.
    \item {\bf Pairwise RAG Comparison} asks models to perform nuanced judgment and contextual understanding.
    \item {\bf Essay Grading} serves as a reference point, requiring semantic understanding where encoders like BERT might already perform well with a reasonable amount of fine-tuning data.
\end{itemize}

These tasks can be framed as  instances of LLM-as-a-Judge evaluation, an area 
where practitioners have identified binary or coarse scoring as a critical bottleneck at scale, finding that it ``collapses under real-world complexity'' by hiding the distinction between nearly correct and entirely wrong responses~\cite{sinha2025diagnostic}. RiR formalizes this observation: producing calibrated continuous judgments requires reasoning that goes beyond what coarse classification demands. Below we identify the appropriate metric for evaluating RiR methods and further describe the tasks above.

\paragraph{Regression Metrics}
\label{sec:metric:ccc}

Normalized Mean Square Error (NMSE) is a common metric for reporting regression performance: $\sum\limits_{i}^n(y_i - \hat{y}_i)^2/\sum\limits_{i}^n(y_i - \bar{y})^2 $,
where $n$ is the size of the dataset, $\hat{y}_i$ is a prediction, $y_i$ the corresponding ground truth value, and $\bar{y}$ is the mean.

\begin{figure*}[t]
    \centering
   \includegraphics[width=\textwidth,]{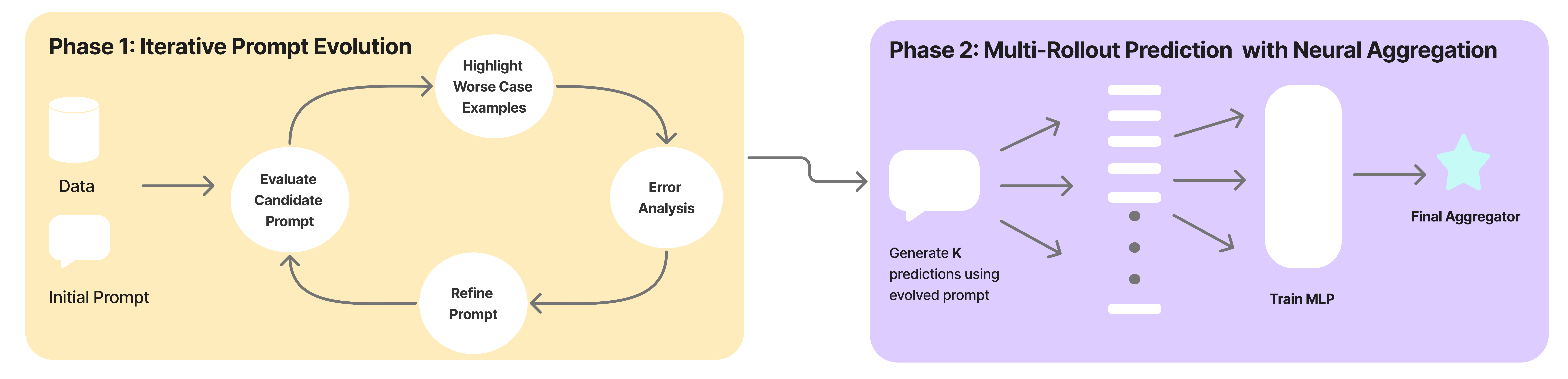} \\
\caption{Phase~1 performs prompt evolution through iterative and \textit{batched} reflection. Given a candidate prompt, we collect rollouts on $n$ samples, divided into training and validation sets. A model instructed to focus on the $\sqrt{n}$ worst-performing examples (selected based on absolute prediction error) analyzes the rollouts on the training samples, in light of the optimization history from previous iterations, and makes proposals that refine the prompt. This cycle continues for a predetermined number of iterations, after which we select the best-performing prompt $P_{\text{best}}$ as evaluated on the validation set , where by best we mean the prompt that led to the highest CCC value. Phase~2 generates multi-rollout predictions by applying $P_{\text{best}}$ and aggregating $K$ stochastic predictions per input and trains a neural aggregator $f_\theta$ on sorted rollouts using a combined  CCC--NMSE loss. Test predictions are obtained by sampling test rollouts and applying the trained aggregator $f_{\theta^{\star}}$.}
\label{fig:mentat}
\end{figure*}

But distance-based metrics are inadequate for typical RiR problems; RiR systems can artificially lower their NMSE simply by avoiding ``risky'' predictions at the extremes. This can be seen in Figure \ref{fig:mentat_v_others} earlier, particularly in comparing the fine-tuned NeoBERT model~\citep{neobert} against detailed (human-crafted) prompting. Following Figure \ref{fig:mentat_v_others}, if we were to rely on NMSE, detailed prompting for \gpt{} would not appear to substantially outperform NeoBERT ($0.81$ vs. $1.01$), and this gap would be even reversed for weaker LLMs. Examining the distribution of predictions reveals that NeoBERT ``hacked'' the loss function by learning a collapsed distribution, while the prompted LLM actually shows substantial signs of ranking the inputs correctly.

This can be captured in a Concordance Correlation Coefficient (CCC) of $0.01$ for NeoBERT versus a CCC of $0.69$ for detailed prompting. We thus suggest the use of the CCC as an additional, and perhaps more appropriate, RiR metric. CCC measures both correlation and agreement, defined as $\frac{2\rho\sigma_y\sigma_{\hat{y}}}{\sigma_y^2 + \sigma_{\hat{y}}^2 + (\mu_y - \mu_{\hat{y}})^2}$, where $\rho$ is the Pearson correlation coefficient between predictions $\hat{y}$ and ground truth $y$, $\sigma_y$ and $\sigma_{\hat{y}}$ are their respective standard deviations, and $\mu_y$ and $\mu_{\hat{y}}$ are their means.
CCC penalizes systematic bias and rewards predictions that maintain the natural variance of the distribution.

\paragraph{Detecting Mathematical Errors}
\label{FME}

We derive a dataset for predicting the \textit{fraction} of a mathematical solution up to the first erroneous reasoning step, given a problem and incorrect solution in LaTeX, from ProcessBench~\citep{processbench}. To effectively do this, a model must systematically reason formally about math steps rather than relying on probabilistic heuristics, but it must also be good at estimating relative lengths and inferring the boundaries of the steps in a calibrated way.

To convert the original classification task into a regression problem, we first filter out problems with correct solutions or final answers. We then merge all solution steps into a single continuous text $T = s_1 \| s_2 \| \cdots \| s_n$ (here $\|$ denotes concatenation). Next, for a solution with error at step $k$, the regression score $R$ is $10 \times (\sum\limits_{i=1}^{k-1} |s_i| + \frac{1}{2}|s_k|)/|T|$ where $|s_i|$ denotes the length of step $i$, and $|T|$ is the total length of the concatenated solution. See an example entry in Appendix \ref{app:examples}. 

\paragraph{Instruction Following}
\label{IF}

We derive a task from the \textsc{WildIFEva} corpus~\citep{wild-data} that targets \emph{instruction-following} in long-form generation. Each example consists of: (i) a user \textit{task} prompt; (ii) a list of atomic requirements (the \textit{decomposition}); (iii) a \textit{model answer} produced by Llama-3.1-8B (zero-shot); and (iv) per-requirement satisfaction scores originally produced by Llama-3.1-70B acting as an automatic judge. The goal is to predict a single continuous label $y \in [0,1]$ that reflects the overall degree to which the answer adheres to the decomposed instructions. More precisely, for each decomposition instance, the judge produced a probability-like score $s_i \in [0,1]$ for each requirement $r_i$, $i=1,\dots,K$. We then use the harmonic mean of these scores as our overall judgment, emphasizing the need to adhere well to all task requirements. To test instruction following, we do not expose the decomposition to NeoBERT or an LLM; instead, they are only given the task and model answer and must infer the overall score.

\paragraph{Pairwise RAG Comparison}
\label{PC}

We derive a dataset for comparing two LLM outputs on a scale from the RAG-QA evaluations~\citep{rag-qa}. Each query $q\in \mathcal{Q}$ has responses $A_1, A_2$ and a target comparative score from $-2$ to $2$ representing the \textit{average} annotation of three human judges, who were instructed to assess response helpfulness, truthfulness, and completeness. Here, positive scores means that $A_1$ is better (and vice versa). This task partially aligns with RiR as judging the outputs and comparing them in light of each query often requires nuanced judgment. 

\paragraph{Essay Grading}
We lastly use an essay grading dataset~\citep{essay}, where each entry contains among other features an essay prompt, a student (grade 8--12) response, associated demographic information, and an overall score between $1$ and $5$. Although Essay Grading is simpler than the rest, it serves as a reference point for the other RiR tasks.

We evaluate these tasks using two proprietary LLMs (\gptf{}, \gpt{}) across three tasks. For Instruction Following, we utilize an open-source model (\gpto{}) for reproducibility and generalization validation.

\section{MENTAT}

\mentat{} combines two simple ideas, depicted in Figure \ref{fig:mentat}: it allows the LLM itself to reflect in batches to incrementally adjust its own prompt, and it aggregates multiple rollouts from the optimized LLM system with a simple trained MLP.

\subsection{Phase 1: Prompt Evolution}

\mentat{}'s first step is to make sure that the LLM prompt reflects both \textit{local} instructions for reasoning about each input and \textit{global} guidance about the distribution of ground-truth scores. Though any approach for prompt optimization can be used here, e.g., MIPRO~\citep{mipro} or GEPA~\citep{gepa}, through preliminary experiments we identified two special properties in RiR tasks that call for different design choices.

First, performing rollouts with powerful \textit{reasoning} models can be expensive and slow, when compared to standard LLMs, for which existing optimizers were built. To remain within the lightweight constraints of typical RiR tasks, a suitable prompt evolution stage would have to minimize \textit{both} the number of rollouts performed with the LLM and the number of \textit{inherently sequential stages} or iterations of optimization. Second, RiR tasks require attention to distributional properties, calibration, variance matching, and avoiding collapse to mean predictions, beyond per-example accuracy. This is because \mentat{}'s aggregation design demonstrates that it can be easy to turn a \textit{well-calibrated} system into one that has low pointwise error, but the reverse is not necessarily true.

This motivates us to test an exceedingly simple reasoning-based technique for optimizing LLM systems that contain a single prompt.\footnote{We leave extending this method to multi-stage LLM programs and conducting an extensive comparison of different prompt optimization strategies to future work.} While batch-based prompt optimization has been extensively explored in prior work~\citep{auto, promptengineer}, we focus on combining it with neural aggregation specifically for regression tasks, using CCC alongside NMSE to guide prompt selection and aggregator training. This simple design is inspired by human prompt engineering practice~\citep{husain_shankar_ai_evals_course_2024}.

Concretely, we proceed in a very small number of sequential iterations (three in our experiments). In each iteration, the work is highly parallelizable: we evaluate the current prompt on a shuffled sample of the training set, and then concatenate all of the rollouts for analysis {\it by the same LLM}. It is then asked to identify systematic errors by analyzing the worst-performing examples and to generate improved instructions. In each iteration, the LLM receives three key inputs: current instructions, performance analysis with detailed error patterns, and a formatted history of previous optimization attempts. This historical context prevents the method from cycling through previously unsuccessful approaches and enables progressive refinement. At the end of this process, the best-performing prompt (via NMSE or CCC) is selected on a separate validation set.

In our evaluation, to stress \mentat{}, we start from a deliberately basic prompt for each task, to reflect a more challenging and informative setting.\footnote{Examples of the basic vs. the detailed prompts used for the four tasks can be found in Appendices \ref{app:basic-prompts} and \ref{app:naive-prompts}), respectively. They differ in the inclusion of detailed procedural steps, calibration guidance, and/or domain-specific heuristics that human experts may decide to include.}
Note also that this iterative prompt evolution follows a single optimization trajectory. In principle, \mentat{} could employ multiple random restarts, which could be parallelized to explore diverse regions of the prompt space. However, we focus on single-trajectory optimization both for computational efficiency and algorithmic simplicity. 

\subsection{Phase 2: Multi-Rollout Generation with Neural Aggregation}

Using the best LLM-discovered prompt from Phase 1, \mentat{} generates multiple independent predictions for each example.
The multi-rollout approach captures the inherent uncertainty in LLM predictions, as each rollout can reason independently, and provides richer signal for the subsequent neural aggregation phase. In practice, we set this to three rollouts per example.

We train a small Multi-Layer Perceptron (MLP) to combine rollout predictions. The aggregator ensures order invariance by sorting rollout predictions, incorporates statistical features (mean, standard deviation, min, max), and is optimized for a combination of the CCC and NMSE loss functions. Overall, this method builds on self-consistency ~\citep{selfconsistency} and best-of-N voting~\citep{stiennon2020learning,snell2024scaling}, but differs by training a lightweight aggregator that learns task-specific weighting of rollout statistics rather than using fixed aggregation rules.

\section{Evaluation}
\label{app:experimental-setup}

In our main experiments, we define two standard baseline approaches for RiR problems: fine-tuning a small Transformer encoder and prompting an LLM, and use these two to understand the relative merits of our method \mentat{} and to develop a series of ablation experiments. Additionally, we compare against~\citet{gepa}, a recent reflective prompt optimization method, to situate \mentat{} relative to modern prompt optimizers.

\subsection{Baseline: Fine-tuning a Transformer Encoder}

We formulate RiR as supervised regression using a 250M-parameter NeoBERT model. The architecture processes minimally formatted text sequences (e.g., combining problem statements with solutions for math errors, augmented with domain-specific prompts).

Inputs are tokenized using NeoBERT's byte-level BPE tokenizer, truncated or padded to 1024 tokens, and passed through the pretrained encoder. The model extracts representations from the [CLS] token, applies dropout regularization ($p=0.2$), and uses linear projection for scalar predictions. The optimization objective minimizes weighted NMSE and CCC using AdamW~\citep{adamW}. This architecture requires only prompt templating beyond standard fine-tuning, with hyperparameters detailed in Appendix~\ref{app:roberta-impl}.

\subsection{Baseline: Prompting a Large Language Model}
We employ Chain-of-Thought style prompting to encourage frozen LLMs to perform explicit reasoning through step-by-step token generation. Our evaluation uses two proprietary models with different reasoning capabilities: \gptf{} (non-reasoning) and \gpt{} (reasoning) across three tasks (Mathematical Error Detection, Pairwise RAG Comparison, and Essay Grading). For the Instruction Following task, we employ \gpto{}, an open-source model, to demonstrate that our method generalizes beyond proprietary systems and to provide more easily reproducible baselines for the community. We note that evaluating MENTAT on even smaller open-source models remains an important direction for validating on-premise deployability. The detailed prompts for all tasks help guide the decomposition of complex inputs and the templates can be found in Appendix \ref{app:naive-prompts}.

This approach is motivated by several practical advantages. Frozen LLMs can act as a unified interface across various natural language tasks, with very little to no training data. This is especially valuable in RiR domains where annotated datasets are often scarce. Utilizing a shared, unified, and amortized infrastructure (i.e., LLM servers) enables us to deploy a single model across many tasks, significantly reducing the computational and financial overhead compared to training multiple specifiable models.

\subsection{Additional Comparisons}
We supplement our main baselines with two additional comparisons that probe different axes of the RiR problem. A state-of-the-art \textit{reasoning reward model} tests whether models explicitly trained for preference judgment transfer to our continuous regression setting, while \textit{reinforcement learning (RL) fine-tuning} allows us to test whether policy gradient methods can effectively optimize for RiR objectives given comparable compute. Each comparison is restricted to a single task: the reasoning reward model to pairwise RAG comparison, where preference modeling is most natural, and RL fine-tuning to instruction following, the task employing an open-source model whose weights can be updated.

\subsubsection{Reasoning Reward Model}

Pairwise preference judgment, the core task underlying reward modeling for RLHF, is a canonical example of reasoning-intensive regression. Recent work on Reasoning Reward Models~\cite{chen2025rmr1} demonstrates that accurate preference judgments require explicit multi-step reasoning: inferring latent evaluation criteria, weighing trade-offs across dimensions (helpfulness, truthfulness, completeness), and grounding judgments in response content rather than surface features. Critically, their findings validate our RiR hypothesis: the bottleneck for preference modeling is \emph{reasoning capacity}, not model scale, as reasoning-enhanced models outperform 70B+ parameter baselines despite being 5$\times$ smaller.

This connection motivates two additions to our evaluation. First, we include a 
recent reasoning reward model as an additional baseline for
the pairwise RAG comparison task, \reasm. Second, since such models produce 
discrete preference labels rather than continuous scores, we extract regression 
targets via differences of log-probability $\hat{y} = 4 \cdot \sigma\bigl(\log p(A) - \log p(B)\bigr) - 2$, where $\sigma$ denotes the sigmoid function and $A$, $B$ are preference tokens. This maps the model's implicit preference strength to our target interval $[-2, 2]$, preserving the continuous nature of judgment confidence rather than discretizing to binary labels.

\subsubsection{RL Fine-Tuning}
\label{sec:rl-finetuning} 
We additionally evaluate RL-based fine-tuning using Group Relative Policy Optimization~\cite{grpo} on the instruction following task. We restrict this comparison to instruction following because it is the only task that employs an open-source model (\texttt{gpt-oss-20b}), whose weights can be updated; the remaining three tasks use proprietary models (GPT-4.1, GPT-5) that are accessible only through frozen inference APIs. We fine-tune a LoRA adapter on \texttt{gpt-oss-20b} via the Tinker API, sampling $k$ rollouts per problem at each training step and computing per-rollout rewards as $r = 1 - |y_{\text{pred}} - y_{\text{true}}|$, where predictions are parsed from the model's chain-of-thought output. Advantages are centered and normalized within each group of rollouts for the same problem, following the standard GRPO formulation, and applied to a PPO-clipped surrogate objective. Other training objectives where experimented with but garnered little benefits (see Appendix \ref{app:ccc-grpo}).

\subsection{Experimental Setup}
Our experimental design evaluates \mentat{} across four reasoning-intensive regression tasks using a structured approach. We test \gptf{} and \gpt{} on three tasks (Mathematical Error Detection, Pairwise RAG Comparison, and Essay Grading), while Instruction Following uses \gpto{} to demonstrate generalization to open-source models.

\begin{table*}[t]
\caption{Performance comparison across Mathematical Error Detection, Pairwise RAG Comparison, and Essay Grading using \gptf{} and \gpt{} as our models. Each entry is the average of three independent runs on a test set of size $750$. Total training sizes are $100$ and $500$ (train/val combined). \textbf{Ablations:} \mentat{} Prompt uses only error-driven prompt refinement on training data {\it starting from a basic prompt}; \mentat{}-Avg shows performance when replacing the trained MLP with averaging. We remark here that NeoBERT obtains an average NMSE and CCC of $0.60$ and $0.66$ respectively, on a training regime of $1500$ ($1000$ training + $500$ validation) on Pairwise RAG Comparison. That is, NeoBERT needs much more data on this task to lead to good performance, but it can be achieved. This table along with additional reporting of standard deviation can be found in Table~\ref{app:tab1-std} in the appendices.}
\centering
\scriptsize
\begin{tabular}{ll cc cc cc cc cc cc}
\toprule
\multirow{3}{*}{\textbf{LM}} & \multirow{3}{*}{\textbf{Method}} 
& \multicolumn{4}{c }{\textbf{Math Errors}} 
& \multicolumn{4}{c }{\textbf{Pairwise RAG}} 
& \multicolumn{4}{c}{\textbf{Essay Grading}} \\
\cmidrule(lr){3-6} \cmidrule(lr){7-10} \cmidrule(lr){11-14}
& & \multicolumn{2}{c }{NMSE $\downarrow$} & \multicolumn{2}{c }{CCC $\uparrow$} 
& \multicolumn{2}{c }{NMSE $\downarrow$} & \multicolumn{2}{c }{CCC $\uparrow$} 
& \multicolumn{2}{c }{NMSE $\downarrow$} & \multicolumn{2}{c}{CCC $\uparrow$} \\
\cmidrule(lr){3-4} \cmidrule(lr){5-6} \cmidrule(lr){7-8} \cmidrule(lr){9-10} \cmidrule(lr){11-12} \cmidrule(lr){13-14}
& & 100 & 500 & 100 & 500 & 100 & 500 & 100 & 500 & 100 & 500 & 100 & 500 \\
\midrule
\multicolumn{13}{c}{\textbf{Main Methods}}\\
\midrule

\midrule
 NeoBERT & Gradient Descent & 1.05  & 1.01 & 0.02 & 0.06 & 1.44 & 1.02 & 0.02  & 0.10 & 1.03 & 0.91 & 0.19 & 0.65 \\
\midrule

GPT-4.1 &  Basic Prompt &
1.59 & 1.59 & 
0.36 & 0.36 &  
2.18 & 2.18 & 
0.47 & 0.47 & 
0.75 & 0.75 & 
0.63 & 0.63\\
& Detailed Prompt & 
1.13 & 1.13 & 
0.52 & 0.52 &  
2.20 & 2.20 &
0.47 & 0.47 & 
0.73 & 0.73 & 
0.65 & 0.65\\
& \mentat$_\text{Basic Prompt}$ & 0.87 & 0.76 & 0.51 & 0.49 & \textbf{0.77} & {\bf 0.80} & \textbf{0.50} & {\bf 0.52} & \textbf{0.54} & \textbf{0.53} & \textbf{0.70} & \textbf{0.68} \\
\midrule

GPT-5 & Basic Prompt &
0.77 & 0.77 & 
0.66 & 0.66 &
2.25 & 2.25 & 
0.35 & 0.35 & 
1.31 & 1.31 & 
0.42 & 0.42 \\
& Detailed Prompt & 0.78 & 0.78 & 0.69 & 0.69 &
2.18 & 2.18 & 
0.31 & 0.31 & 
1.53 & 1.53 & 
0.40 & 0.40\\
& \mentat$_\text{Basic Prompt}$ & {\bf 0.52} & {\bf 0.42} & {\bf 0.72} & {\bf 0.78} & 1.07 & 0.93 & 0.36 & 0.33 & 0.64 & 0.67 & 0.59 & 0.55 \\

\midrule
\multicolumn{13}{c}{\textbf{Ablations}}\\
\midrule

GPT-4.1 & \mentat{} Prompt & 1.39 & 1.29 & 0.45 & 0.48 & 2.00 & 1.69 & 0.45 & 0.48 & 0.61 & 0.71 & 0.68 & 0.66 \\
& \mentat-Avg & 1.00 & 1.01 & 0.52 & 0.52 & 1.82 & 1.48 & 0.48 & 0.51 & 0.57 & 0.63 & 0.69 & {\bf 0.68} \\
& \gepa & 1.04 & 1.01 & 0.49  & 0.54 & 2.16  & 2.40 & 0.44  &  0.43 & 0.79  & 0.81 & 0.63 &  0.63 \\

GPT-5 & \mentat{} Prompt & 0.66 & 0.58 & 0.66 & 0.72 & 1.43 & 1.95 & 0.33 & 0.30 & 0.74 & 0.70 & 0.57 & 0.54 \\
& \mentat-Avg & 0.59 & 0.51 & 0.68 & 0.75 & 1.31 & 1.83 & 0.35 & 0.32 & 0.69 & 0.67 & 0.57 & 0.55 \\
& \gepa & 0.78  & 0.63  & 0.68  & 0.69  & 2.48 & 2.29  & 0.28 & 0.28 & 1.01  & 1.01  & 0.42 & 0.44  \\
\bottomrule
\end{tabular}
\label{tab:combined-results}
\end{table*}

For the three tasks using proprietary models, we employ $750$ test examples with results averaged across three independent runs. We evaluate under two training configurations ($100$ and $500$ samples) that reflect real-world data constraints typical in ad-hoc RiR applications. For prompt optimization methods (including  \mentat{}), we use balanced train/validation splits of $50+50$ and $250+250$ samples; Phase 1 uses these for prompt evolution, and Phase 2 generates 3 rollouts per training sample for MLP training. For NeoBERT fine-tuning, we employ training-heavy splits of $50+50$ and $350+150$ samples to leverage the model's supervised learning capabilities.

For Instruction Following, we use a single configuration with $500$ training, $500$ validation, and $2000$ test samples, reflecting the different data availability typical for this fundamental capability assessment. This larger test set enables more robust evaluation of the nuanced instruction-adherence requirements.

This experimental structure allows us to assess \mentat{}'s effectiveness across different model capabilities (reasoning vs. non-reasoning), data regimes (limited vs. moderate training data), and model accessibility (proprietary vs. open-source), providing comprehensive validation of our approach for practical RiR applications.

\subsubsection{Computation Cost}
\mentat{}'s  computational costs comprise two phases. At inference time, each prediction requires 3 rollouts, resulting in $3\times$ token cost compared to  single-pass prompting. However, all rollouts can be generated in parallel, so wall-clock latency remains approximately equivalent to a single rollout in parallelized deployment scenarios.

During optimization (Phase 1), \mentat{} uses a fixed $3$-iteration design. Each iteration evaluates the current prompt on all $n=250$ training and $n=250$ validation samples (parallelizable) and performs one reflection call analyzing the $\sqrt{250} \approx 16$ worst-performing examples. Including the initial baseline evaluation, this totals approximately $2{,}003$ ($2 \times 4 \times 250 + 3$) LLM calls across $4$ sequential stages. \gepa{}'s ``light'' configuration converges after an average of $23$ sequential iterations (ranging $15$--$34$ across runs). \gepa{}'s evolutionary search thus requires approximately $8\times$ more sequential rounds than \mentat{}'s fixed design, providing \mentat{} a substantial wall-clock advantage in parallelized deployments, though total token consumption may differ.

Phase 2 generates $3$ stochastic rollouts per training and validation example, adding $3 \times 2n = 1{,}500$ LLM calls, though predictions from Phase~1's best iteration can serve as one rollout, reducing this to $2 \times 2n = 1{,}000$. The MLP aggregator itself has negligible cost, containing only $8$ hidden units and training on $750$ rollout vectors ($250$ samples $\times 3$ rollouts each).

\paragraph{Instruction Following Parameters} For the instruction following task, we select GRPO hyperparameters to match \mentat{}'s optimization budget. \mentat{}'s Phase~1 (prompt evolution) requires $4{,}003$ LLM calls across four sequential stages: one baseline evaluation and three refinement iterations, each evaluating $500$ training and $500$ validation examples plus one reflection call. Phase~2 adds $3$ stochastic rollouts per training and validation example, contributing an additional $2{,}000$--$3{,}000$ calls depending on whether Phase~1 predictions are reused.

This yields an optimization budget of approximately $6{,}000$--$7{,}000$ LLM calls (excluding test evaluation). To match this budget, we set batch size $b=12$, group size $k=8$, and train for $50$ steps, giving $12 \times 8 \times 50 = 4{,}800$ training rollouts. We evaluate on the full $500$-example validation set at steps $20$, $40$, and $49$ (final), adding $3 \times 500 = 1{,}500$ validation rollouts, for a total optimization budget of $6{,}300$ LLM calls, closely matching \mentat{}'s compute envelope while allowing a direct methodological comparison.

\subsection{Results}

\begin{figure*}[h]
\centering
\includegraphics[width=.97\textwidth]{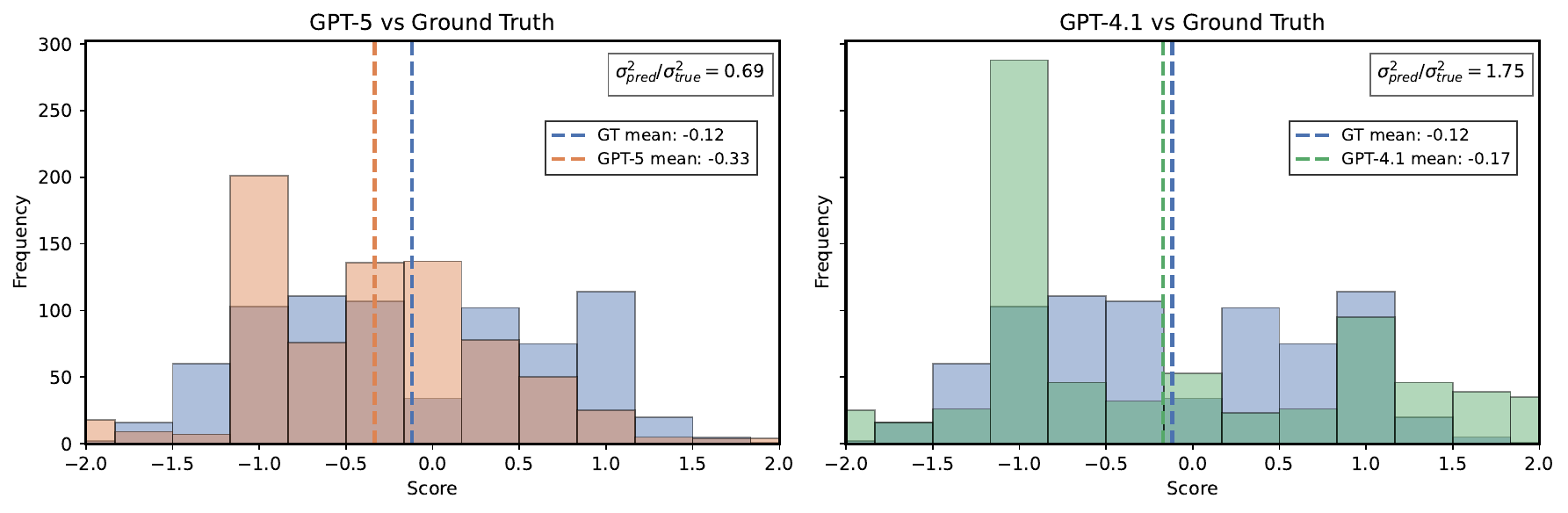}
\caption{Pairwise RAG distributions on the mean of three rollouts vs. ground truth after the prompt evolution process. \gpt{} (left) is center-seeking and under-dispersed; \gptf{} (right) stays closer to the empirical mean and exhibits greater spread. This behavior leads to higher CCC for \gptf{}.}
\label{fig:analysis-rag}
\end{figure*}

Our main evaluation results are reported in Table \ref{tab:combined-results} and \ref{IF-table}, demonstrating significant performance variations across methods and tasks. The results reveal distinct patterns in how different approaches handle reasoning-intensive regression problems, with \mentat{} consistently outperforming baseline methods across most configurations. Beyond aggregate metrics, we analyze failure modes across methods: NeoBERT's distribution collapse (\ref{fig:mentat_v_others}), GPT-5's center-seeking behavior on pairwise RAG (\ref{fig:analysis-rag}), and systematic quantization patterns in LLM outputs (Appendix \ref{quantization}). We provide additional per-task qualitative error analysis in Appendix \ref{app:failure-modes}.

\begin{table}[t]
\caption{Performance on the Instruction Following task using the \gpto{} model. Each entry is the average of three independent runs on a test set of size 2000. Total training configuration uses 500 training and 500 validation samples. Ablations: \mentat{} Prompt uses only error-driven prompt refinement on training data; \mentat{}-Avg shows performance when replacing the trained MLP with averaging. The subscripts,  basic prompt and detailed prompt, are what we use an the initial prompt in the \mentat{} framework. Moreover, values within the parenthesis represent standard deviations.}
\centering
\scriptsize
\begin{tabular}{ll cc cc}
\toprule
\multirow{3}{*}{\textbf{LM}} & \multirow{3}{*}{\textbf{Method}} 
& \multicolumn{4}{c}{\textbf{Instruction Following}} \\
\cmidrule(lr){3-6}
& &\multicolumn{1}{c }{NMSE $\downarrow$} & \multicolumn{1}{c }{CCC $\uparrow$}  \\
\midrule
 NeoBERT & Gradient Descent & 1.08 (0.07) & 0.36 (0.04) \\
\midrule
GPT-OSS-20B &  RL Fine-Tuning &  
1.51 (0.03) &0.37 (0.02)  \\
 \midrule
GPT-OSS-20B &  Basic Prompt &  1.18 (0.00)
 & 0.32 (0.00) \\
& Detailed Prompt & 1.16 (0.00)
 & 0.33 (0.00)\\
& \mentat$_\text{Basic Prompt}$ & 0.95 (0.09)  
 & 0.42 (0.01) \\
& \mentat$_\text{Detailed Prompt}$ & \textbf{0.90 (0.04)}
 & 0.43 (0.00)  \\  
\midrule
\multicolumn{6}{c}{\textbf{Ablations}}\\
\midrule
GPT-OSS-20B &  \mentat{}$_\text{Basic Prompt}$ Prompt & 1.25 (0.05)
 & 0.35 (0.01)  \\
& \mentat$_\text{Basic Prompt}$-Avg & 1.06 (0.04)  
 & 0.38 (0.02) \\
& \mentat{}$_\text{Detailed Prompt}$ Prompt & 1.24 (0.13)
 &  0.36 (0.01) \\
& \mentat$_\text{Detailed Prompt}$-Avg & 1.09 (0.06) 
 & 0.39 (0.02) \\
 & GEPA & 1.06 (0.02) & \textbf{0.46 (0.01)} \\
\end{tabular}
\label{IF-table}
\end{table}

\paragraph{Mathematical Error Detection Performance} 

On this task, fine-tuning NeoBERT achieves near-zero CCC scores across both training configurations and effectively collapsing to mean predictions as shown in Figure \ref{fig:mentat_v_others}. In contrast, LLM-based approaches demonstrated substantial reasoning capabilities. \gptf{} with detailed prompting achieved CCC scores of $0.52$ (100-sample training) and maintained this performance at 500 samples. However, \mentat{} with \gptf{} showed only modest improvements, reaching CCC scores of $0.51$ (100 samples) and $0.49$ (500 samples), representing approximately stable performance with slight variation. We hypothesize that \gptf{}'s limited reasoning capabilities on this reasoning-intensive task made it difficult to understand its own errors and thus improve.

The most dramatic improvements can be seen with \gpt{}. While detailed prompting with \gpt{} achieved strong baseline performance (CCC: 0.69, NMSE: 0.78), \mentat{} with \gpt{} delivered substantial enhancements. In the 100-sample training regime, CCC improved by $4.3\%$, while NMSE improved by $33.3\%$. In the 500-sample training regime, CCC improved by $13\%$, while NMSE improved by $46.2\%$. 
These results indicate that \mentat{}'s iterative prompt refinement and neural aggregation effectively leverage \gpt{}'s reasoning capabilities while addressing the precision limitations inherent in direct LLM numerical prediction.

\paragraph{Instruction Following Performance}
For the instruction following RiR task, the NeoBERT model achieved modest performance (CCC: $0.36$, NMSE: $1.08$), while the \gpto{} model with basic and detailed prompting showed similar limitations (CCC: $0.32$--$0.33$, NMSE: $1.16$--$1.18$). RL fine-tuning via GRPO improved CCC to $0.37$ over frozen prompting, though at the cost of degraded calibration (NMSE: $1.51$). The elevated NMSE despite improved CCC suggests that RL fine-tuning sharpens relative discrimination between examples at the cost of absolute calibration. We emphasize that this comparison is conducted under matched compute budgets; GRPO with substantially greater rollout budgets could potentially close or reverse this gap, consistent with the scaling properties documented by~\citet{gepa}.

\mentat{} demonstrated improvements across both initialization strategies, achieving CCC of $0.42$--$0.43$ and NMSE of $0.90$--$0.95$. Notably, instruction following is the one task where \gepa{} surpasses \mentat{} in correlation (CCC $0.46$ vs $0.43$), though \mentat{} retains a clear calibration advantage (NMSE $0.90$ vs $1.06$). We attribute \gepa{}'s stronger correlation to the nature of the task: instruction following scores exhibit high variance and limited systematic structure, favoring \gepa{}'s unconstrained evolutionary search over \mentat{}'s fixed $3$-iteration design. The ablation results show that neural aggregation improves calibration (NMSE) over prompt evolution alone across both initialization strategies, 
though \gepa{}'s stronger CCC suggests that \mentat{}'s fixed-iteration 
prompt evolution may underexplore the prompt space on this task, limiting 
the quality of the representations passed to the aggregator.

\paragraph{Pairwise RAG Comparison Performance}
\label{weirdness-of-gpt5}

On the pairwise RAG comparison task, fine-tuning NeoBERT achieved very low CCC scores while appearing competitive on the NMSE metric by ``hacking'' the distribution. Surprisingly, \gptf{} demonstrated superior performance compared to \gpt{} on this task, in sharp contrast with the general trend observed in mathematical error detection. Detailed prompting with \gptf{} achieved CCC scores of $0.47$ across both training configurations, while \gpt{} detailed prompting resulted in lower CCC scores of $0.31$. 

Unlike math errors, instruction following, and essay grading tasks, correct decisions on the pairwise RAG benchmark often hinge on a few salient cues and short justifications. With chain-of-thought scaffolds on this task, we observe that \gpt{} systematically ``overthinks,'' resulting in predictions that concentrate near the center ($0$ on the $[-2,2]$ margin) rather than faithfully spreading across the empirical label distribution. As shown in Figure~\ref{fig:analysis-rag}, its variance is under-dispersed relative to ground truth, with more than half of examples yielding identical rollouts across three samples. Rollout correlations are very high, and the final numbers fall on a coarse grid (e.g., $\{-1,-\tfrac{1}{3},0,\tfrac{1}{3}\}$), all consistent with hedging.

By contrast, \gptf{} produces short, decisive judgments that remain closer to the dataset mean with greater spread and more frequent use of the extremes. Although \gptf{} rollouts are also correlated, the resulting distribution retains enough variance and calibration to yield substantially higher CCC. For pairwise RAG, \gpt{} tends toward the center and compresses its numeric range, degrading distributional fidelity (and thus CCC) even when NMSE remains similar.

We hypothesize that \gptf{}'s superior performance on pairwise RAG comparison aligns with recent findings that large reasoning models often underperform on simpler tasks~\citep{illusionthinking}. These models initially find correct solutions but continue reasoning toward incorrect answers, suggesting that excessive sophisticated reasoning can sometimes be counterproductive. This hypothesis is supported by our observation that over half of \gpt{}'s examples yield identical rollouts across three samples, with final scores clustering on a coarse grid rather than reflecting the task's inherent variance.

 \reasm{} achieves an NMSE of 5.66 and CCC of 0.15, substantially underperforming even basic GPT-5 prompting (NMSE 2.25, CCC 0.35). Furthermore, its binary classification accuracy of 55.7\% is only marginally above chance, indicating that the model struggles not only with regression magnitude but also with preference direction. We attribute this poor performance to a fundamental mismatch between RM-R1's training objective, selecting a binary winner, and the RiR requirement of producing calibrated, fine-grained scores.  We adapted \reasm{} for regression by extracting the log-probabilities over the binary preference tokens and mapping them to continuous scores. However, these probabilities reflect the model's confidence in a discrete choice rather than its assessment of preference magnitude. By the time \reasm{} commits to generating its answer token, it has already reasoned through its Chain-of-Rubrics and reached a binary conclusion; the resulting probability distribution is highly concentrated (typically >0.95), providing little signal about whether the preferred response is ``slightly better" versus ``much better." This result underscores that even reasoning-augmented reward models, despite their success on preference benchmarks like RewardBench, do not naturally generalize to reasoning-intensive regression settings.
\paragraph{Essay Grading Performance}

Essay grading represented the least complex reasoning-intensive task, with NeoBERT achieving reasonable performance that improved substantially with additional training data. This aligns with the task's characterization as requiring primarily semantic understanding rather than deep sequential reasoning. \gptf{} achieved strong baseline performance with detailed prompting (CCC: 0.65, NMSE: 0.73), while \mentat{} provided meaningful improvements. In the 100-sample training regime, CCC improved by $7.7\%$ and NMSE improved by $26.0\%$ compared to detailed prompting. In the 500-sample training regime, CCC improved by $4.6\%$ and NMSE improved by $27.4\%$. Notably, \gpt{} performance on essay grading showed surprisingly poor concordance compared to \gptf{}, supporting the hypothesis from Section \ref{weirdness-of-gpt5} that sophisticated reasoning models may over-deliberate on simpler tasks.

\section{Conclusion}

We investigated \textit{reasoning-intensive regression} (RiR). Our empirical findings reveal tension: prompting leverages LLMs' reasoning capabilities but produces quantized, imprecise outputs, while supervised fine-tuning for regression can often collapse without learning the task. We proposed \mentat{}, a simple method that suggests that hybrid approaches may help address this tension through iteratively optimizing the prompts via batched error analysis combined with neural aggregation, achieving consistent improvements across several different RiR tasks. 

However, our work opens several rich avenues for future research. The RiR framework we establish creates opportunities to more extensively evaluate sophisticated RL and prompt optimization techniques and develop RiR-adapted regression-aware fine-tuning methods~\citep{regression-aware,TRACT}. Extending the benchmark beyond its current four tasks, particularly to domains such as clinical scoring, financial risk assessment, and code review, is a primary direction for future work, and would help clarify where the boundary between Level 2 and Level 3 regression lies empirically. Similarly, our open-source experiments currently use \gpto{} and validating MENTAT on even smaller open-source models would strengthen the case for on-premise deployment in regulated settings such as finance or healthcare, where proprietary API access may be restricted. Moreover, our lightweight constraint focus also motivates exploring the efficiency-performance trade-offs in reasoning-intensive tasks. While reinforcement learning methods like Group Relative Policy Optimization~\citep{grpo} require thousands of rollouts that exceed the lightweight compute budgets typical of ad-hoc RiR deployment, our benchmark provides a testbed for developing more efficient alternatives as RiR datasets scale. Similarly, \mentat{}'s $3\times$ inference cost increase highlights the need for systematic cost-benefit analysis across deployment scenarios, opening questions about adaptive rollout strategies and inference-time optimization that our tasks can help address. Lastly, we acknowledge the use of light assistance from generative AI tools, with all outputs reviewed and edited by the authors, in the preparation of portions of this paper.

\subsection*{Acknowledgments}

This work used Expanse GPU at the San Diego Supercomputer Center (SDSC) through allocation CIS250733 from the Advanced Cyberinfrastructure Coordination Ecosystem: Services \& Support (ACCESS; \citealt{ACCESS_PEARC23}) program, which is supported by U.S. National Science Foundation grants $\#2138259$, $\#2138286$, $\#2138307$, $\#2137603$, and $\#2138296$. This research was partly supported by Laude Institute.

\bibliographystyle{ACM-Reference-Format}
\bibliography{custom}
\appendix
\section{Extended Related Work}
\label{related-works}

This appendix presents more expansive related work besides those covered in the main sections.

\textbf{Ensemble Learning.} Ensemble learning combines several {\it individual} models to obtain better performance ~\citep{ensemblingreview}. Classical methods include bagging, boostrapping, as well as stacking ~\citep{bagging, boosting, stacked, stacked-regression}. General methods include negative correlation learning, explicit and implicit ensembles, and homogeneous and heterogeneous ensembles ~\citep{ncl, explicit, homo}. More recent ensembling approaches for LLMs include \llmblender \text{} which seeks to pairwise compare from a set of $N$ different LLMs to discern subtle differences in output, then merges the top $K$ ranked outputs ~\citep{llmblender}. DeePEn ~\citep{deepen} is an ensembling method in which probability distributions from individual LLMs are translated into a ``relative representation'' space (to bypass the vocabulary discrepancies), making aggregation possible. There are many recent works on fusion methods ~\citep{URG,specfuse, packllm,processlevelensem,replacingjudgesjuries}. ~\citet{fusion-comp} propose a fusion-of-experts method which fuses outputs of multiple (expert) models with {\it complementary} knowledge of the data distribution and casts it as a supervised learning problem. {\it Prompt} ensembling has also had great success in improving task accuracy~\citep{cape, boostingprompts, zeroshotprompting,dense,MoRE,askanything,diverse} along with using Recursive Feature Machines (RFMs) for feature learning and aggregation for the steering of LLMs~\citep{beaglehole2025aggregateconquerdetectingsteering}. 

\textbf{Routing.} Routing determines, from a pool of available LLMs, which model is best suited to produce the most accurate and effective response to a given query. Recent work includes \routellm \text{} ~\citep{routellm}, a framework for query routing between ``strong'' and ``weak'' LLMs and \zooter \text{} ~\citep{zooter}, a reward-guided routing approach that distills rewards from training queries into a routing function, enabling precise allocation of each query to the LLM with the relevant expertise.

\textbf{Mixture-of-Experts.} Mixture-of-Experts (MoEs) is a framework in architecture design, in which multiple specialized sub-models (``experts'') handle different parts of the input space ~\citep{moe, moe-jacobs, moe-shazeer}. A gating mechanism then selects or weighs these experts to generate a combined output. Recent work has sought to extend MoEs to LLMs, where several MLP experts are added after each multi-head self-attention module in the Transformer encoder and decoder blocks ~\citep{moe-fredus, moe-palm, moe-fine-tuning, switchhead}. MoEs applications in LLMs have demonstrated demonstrated the ability to increase model size without a proportional rise in computational complexity, largely due to MoEs' inherently sparse computations ~\citep{ICML2024_MoE}. Recently, the  mixture-of-agents~\citep{moa} architecture has been proposed, in which multiple LLMs are stacked into sequential layers. Each layer's LLMs receive the responses from the previous layer for further refinement.

\textbf{Natural Language Regression.} The two common approaches to solving natural language regression using decoder-based LLMs includes {\it autoregressive regression} ~\citep{fromwords, regression-aware, anonymous2024better, forecasters, goat} and {\it predictive head} ~\citep{rankt5, devil}. The former directly predicts the numerical target as text (e.g., predict $112$ by predicting the tokens `1', `1', and `2'). The latter approach learns a separate head on encoded inputs.

Currently, work on advancing regression tends to focus on non-reasoning classical feature-based regression tasks, this includes OmniPred~\citep{omnipred} which introduces a framework for training language models as universal end-to-end regressors. They train a 200M parameter T5 encoder-decoder for the specific task of {\it classical} regression. Complementarily, ~\citet{embedthenregress} introduces an ``embed-then-regress" framework that leverages pre-trained language models' string embedding capabilities to map arbitrary text inputs into fixed-dimensional vectors for downstream regression.

Fine-tuning large language models (LLMs) represents a potential approach for RiR, but recent work \citep{regression-aware,anonymous2024better,TRACT} studying conventional regression problems, generally without any reasoning, demonstrates that decoder-only Transformers face fundamental optimization challenges for regression tasks due to the misalignment between cross-entropy loss (optimized for classification) and regression objectives. Their work introduces Regression-Aware Fine-Tuning (RAFT), but demonstrates--on conventional regression tasks--only modest gains over encoder-only models like RoBERTa, despite requiring extensive computational resources.

Other recent work has explored specific language-oriented regression tasks that involve reasoning, particularly for reward models in particular ~\citep{GRM, critiqueoutloud}. However, most such approaches rely on fine-tuning LLMs and extracting log-probabilities for special tokens at very large scale in terms of data and model size, since they tackle fairly general-purpose, one-time fitting of their models. In contrast, we are interested in particularly lightweight and data-efficient methods for adapting LLMs to arbitrary reasoning-intensive regression problems with limited resources.

\newpage

\section{Numerical Output Quantization in Large Language Models}
\label{quantization}

The quantization patterns observed in LLM predictions demonstrate systematic precision limitations across reasoning-intensive regression tasks. Analysis of the test set per model on the math errors task reveals that \gptf{} exhibits $63.1\%$ clustering at $.00/.50$ decimal endings, while \gpt{} shows $86.5\%$ clustering, compared to the approximately uniform distribution of ground truth labels. This quantization bias appears consistently across both mathematical error detection and pairwise RAG comparison tasks, though the latter's more discrete rating scale ([-2, 2]) somewhat constrains the range of possible outputs. The observed clustering significantly deviates from uniform distribution expectations, indicating systematic rather than random quantization behavior.

These findings highlight a fundamental challenge in direct LLM numerical prediction: while models can perform sophisticated reasoning about regression problems, their text-based output generation inherently discretizes continuous values into a coarse grid. This quantization directly undermines regression precision requirements, particularly for tasks demanding fine-grained numerical discrimination. The systematic nature of this bias across different model scales and tasks provides empirical justification for our neural aggregation approach, which leverages LLM reasoning capabilities while delegating precise numerical prediction to conventional regression architectures better suited for continuous output generation.

\begin{figure*}[h]
\centering
\includegraphics[width=\textwidth]{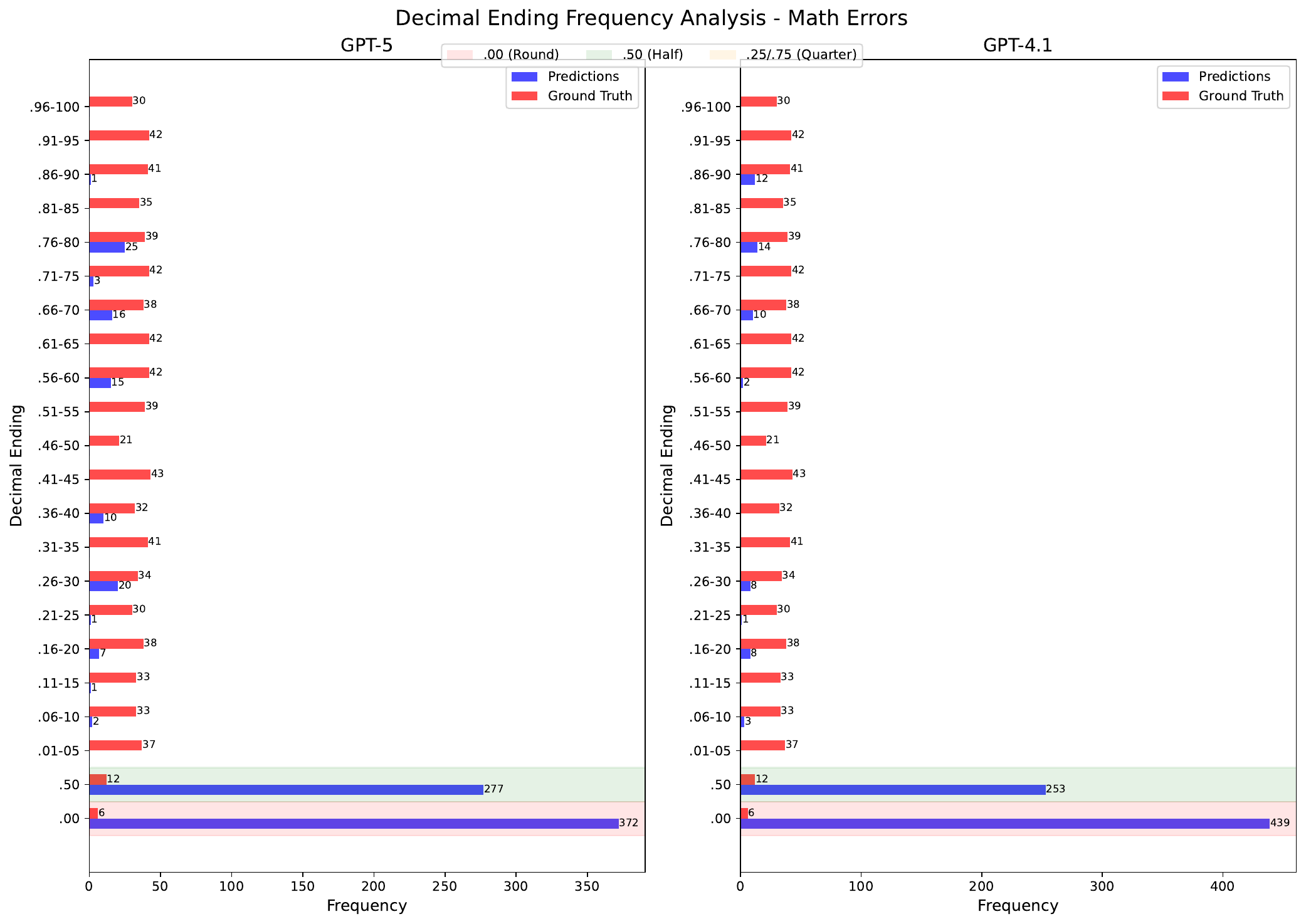}
\caption{Distribution of decimal endings in LLM numerical predictions versus ground truth labels for mathematical error detection task ($n=750$ per distribution). \gptf{} predictions show $63.1\%$ clustering at $.00/.50$ endings ($439+253$ out of $750$ valid predictions), while \gpt{} shows $86.5\%$ clustering ($277+372$ out of $750$ valid predictions). Ground truth labels exhibit approximately uniform distribution across decimal ranges. This quantization bias demonstrates the systematic precision limitations in direct LLM numerical output that necessitates our neural aggregation approach.}
\label{fig:math-errors-decimal-ending}
\end{figure*}

\begin{figure*}[h]
\centering
\includegraphics[width=\textwidth]{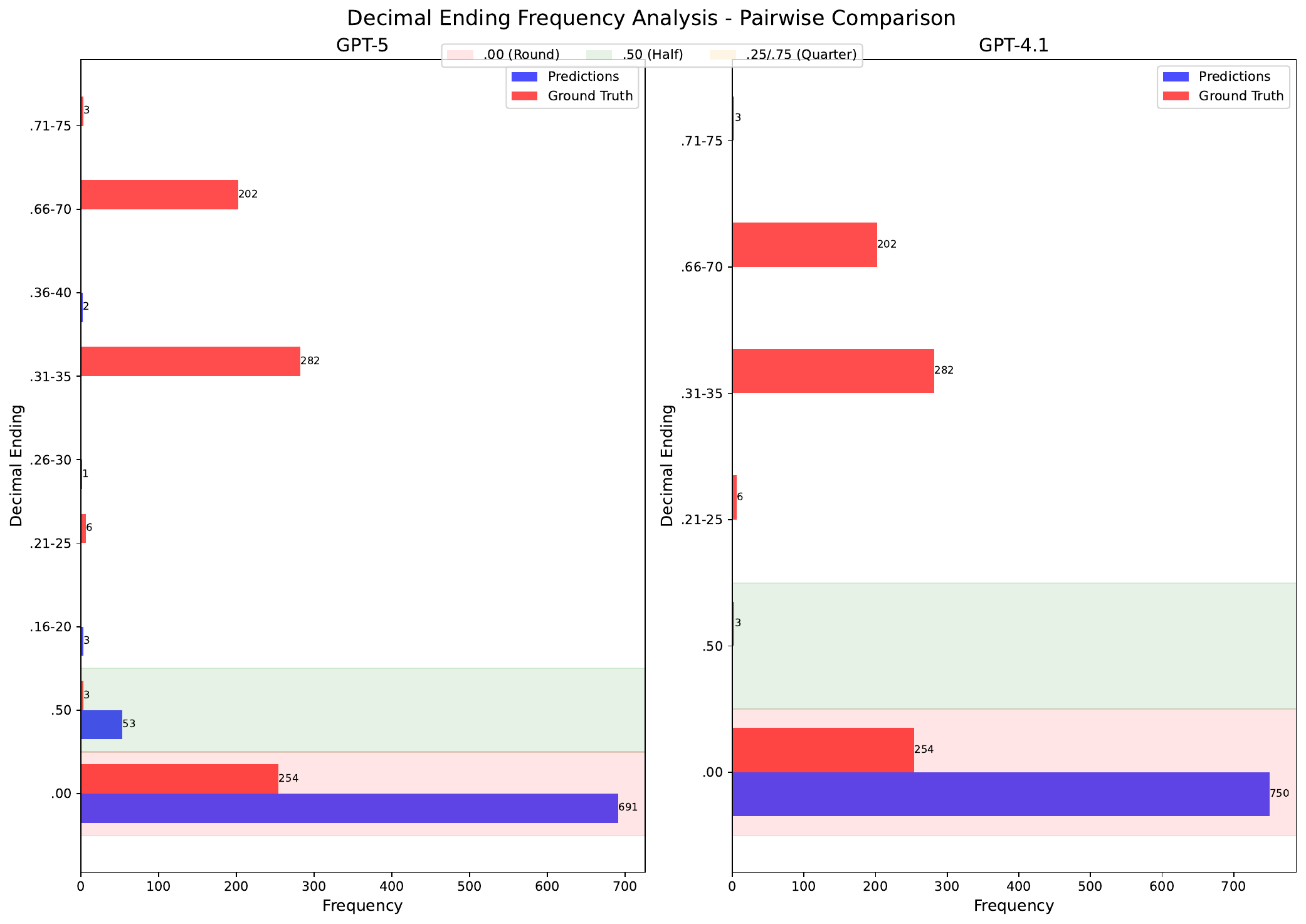}
\caption{Distribution of decimal endings in LLM numerical predictions versus ground truth labels for pairwise RAG comparison task ($n=750$ per distribution). \gptf{} predictions show $100\%$ clustering at $.00$ endings, while \gpt{} shows $99.2\%$ clustering at $.00/.50$ endings ($691+53$ out of $750$ predictions). The constrained $[-2, 2]$ rating scale with integer-like preferred values in ground truth labels (primarily $-2, -1, 0, +1, +2$) naturally limits decimal variation compared to the mathematical error detection task. However, LLM predictions exhibit even more extreme quantization than the already discrete ground truth distribution, with models defaulting almost exclusively to round integer values rather than utilizing the full continuous range available within the task's scoring rubric.}
\label{fig:pc-decimal-ending}

\end{figure*}

\newpage
\section{Failure Modes of RiR Tasks}
\label{app:failure-modes}

\subsection{Mathematical Error Detection}
\begin{figure*}[h!]
    \caption{Absolute prediction error versus solution length for the math-error regression task using \mentat{} with \gpt{} as the model. Across two runs with differing prompts, errors show no meaningful dependence on solution length (average Pearson correlation $\rho = 0.05$). High- and low-error examples occur at all chain-of-thought lengths, indicating that performance is not primarily driven by verbosity but by deeper semantic factors of the problems themselves}
    \centering
    \includegraphics[width=0.40\textwidth]{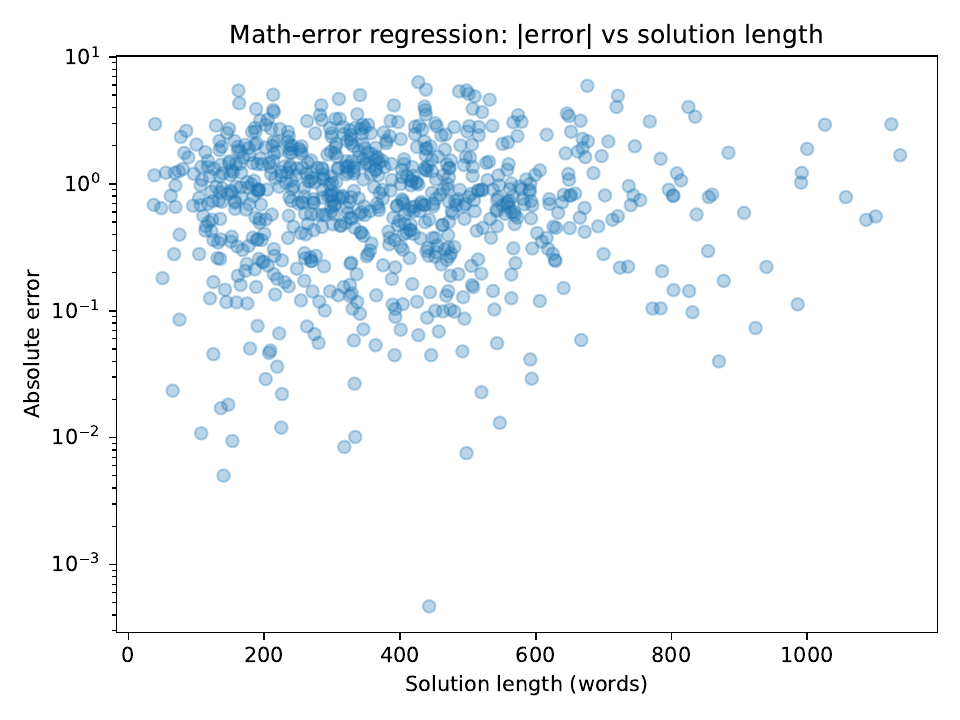}
    \includegraphics[width=0.40\textwidth]{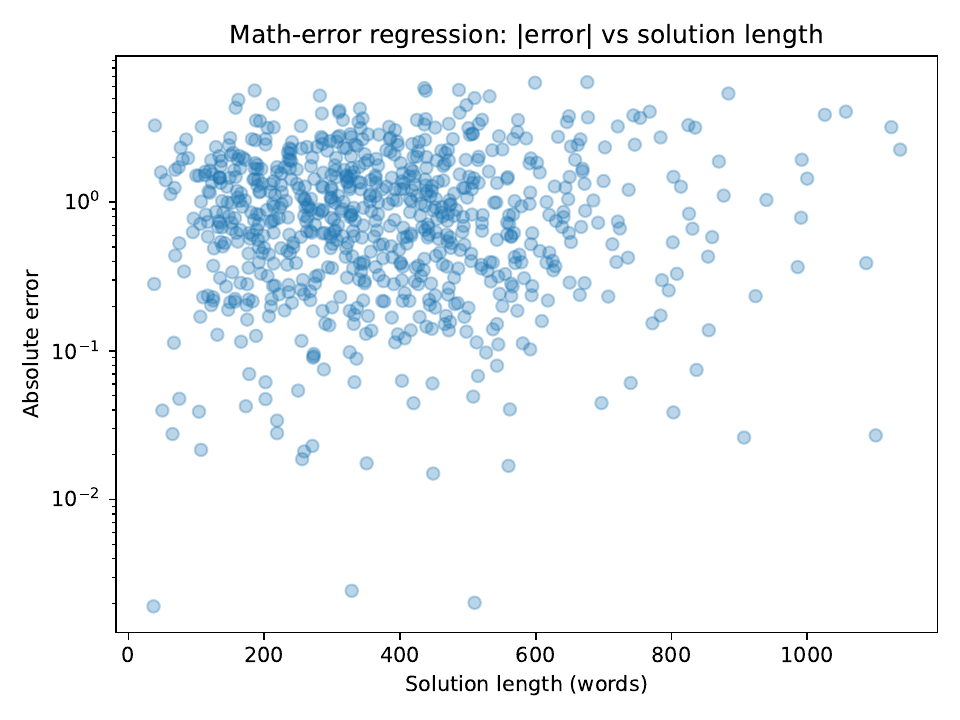}
\label{app:solution-length-analysis}
\end{figure*}

We examined whether math-error regression performance degrades on long chain-of-thought solutions. As shown in Figure \ref{app:solution-length-analysis}, absolute prediction error shows no strong dependence on solution length (average $\rho$ of $0.05$). Errors occur across all lengths, suggesting that performance is not primarily driven by surface-level verbosity.

\begin{figure*}[t]
    \centering
   \includegraphics[width=.84\textwidth]{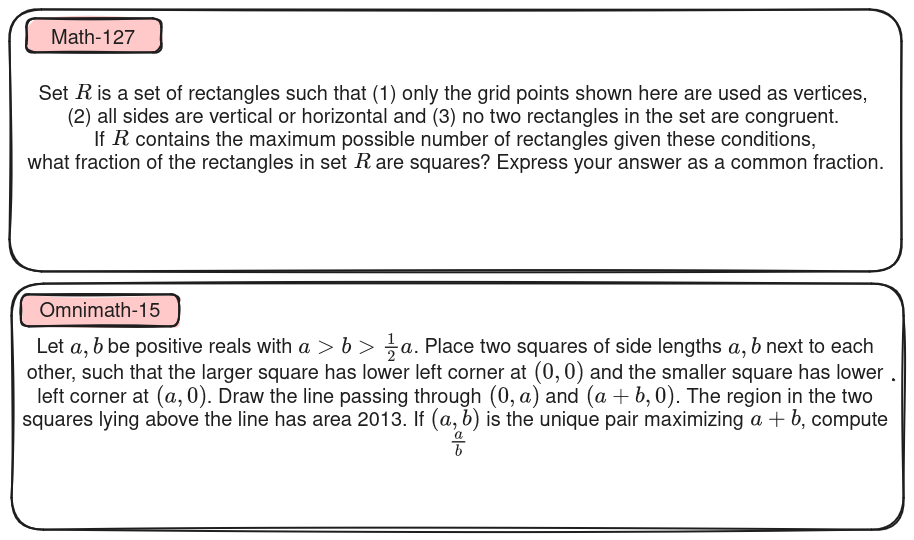} \\
\caption{Two math errors problems with large absolute predictions errors ($6.31$ for Math127 and $6.41$ for omnimath15). Both problems rely on heavy geometric intuition to solve.}
\label{app:geoemtry-cases}
\end{figure*}

We instead found through qualitative analysis of high-error cases that their is a distinct concentration in geometry and spatial-reasoning problems (e.g., grid-rectangle enumeration, line–region intersection). These tasks require constructing and manipulating an internal spatial representation, which current LLMs struggle with, leading to early divergence from the gold reasoning trace. This is in line with current finding on the difficult LLMs face with respect to geometric reasoning~\citep{geometryllm}. We present two problems with very large prediction errors in Figure \ref{app:geoemtry-cases}.

\subsection{Pairwise RAG Comparison}
\begin{figure*}[h!]
    \centering
    \includegraphics[width=0.40\textwidth]{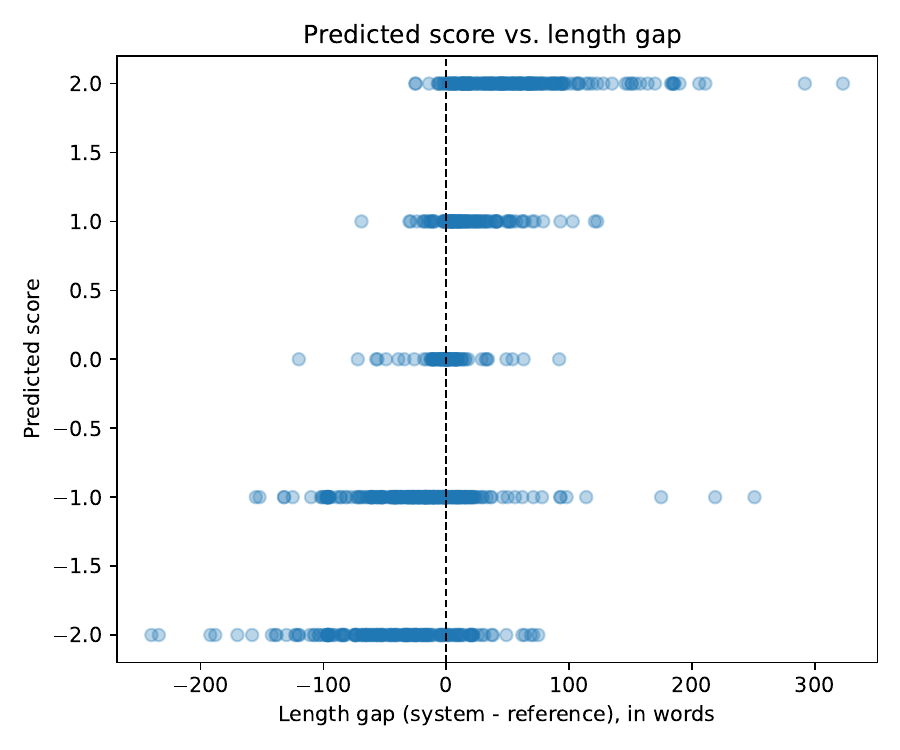}
    \includegraphics[width=0.40\textwidth]{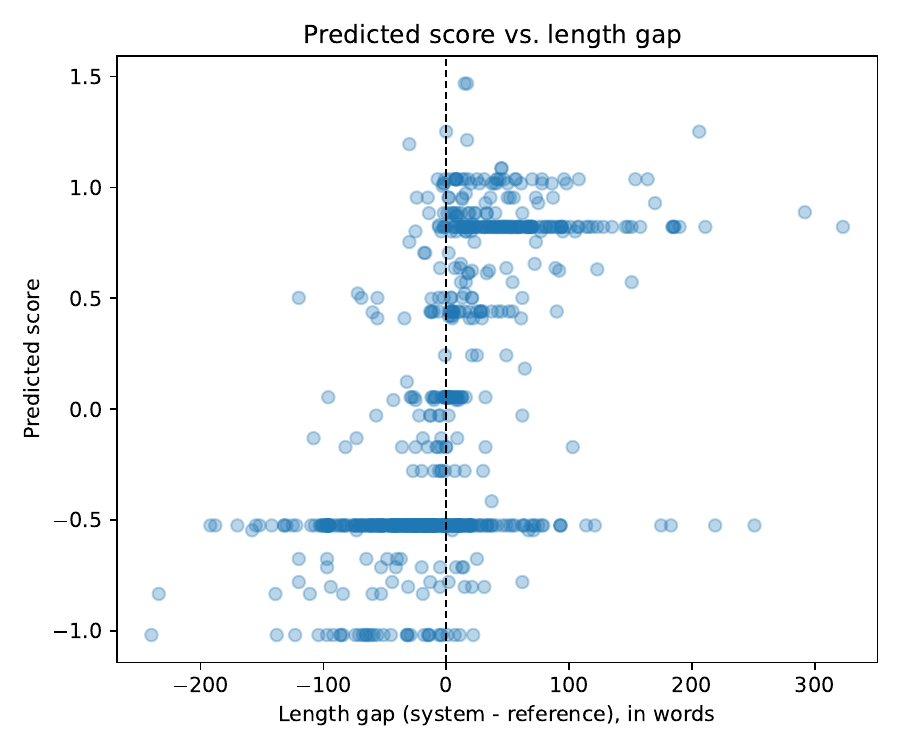}
    \caption{Predicted score versus length gap ($\Delta$ length = sys\_len - ref\_len) for pairwise RAG comparison  evaluation. Left:  Detailed prompting exhibits a strong monotonic length bias, strongly preferring longer system responses ($\rho = 0.617$). Right: \mentat{} reduces this effect ($\rho = 0.551$), producing a flatter trend around $\Delta$ length $= 0$. Human annotations themselves display notable length correlation ($\rho = 0.332$), explaining why neither method can eliminate the bias entirely.}
    \label{app:ref-sys-length-analysis}

\end{figure*}

We analyze length bias in pairwise RAG comparison scoring by measuring how predicted preference scores vary with the length gap between the system and reference responses ($\Delta$ length = sys\_len − ref\_len). As shown in Figure \ref{app:ref-sys-length-analysis}, human annotation scores already exhibit a l correlation with response length ($\rho = 0.332$), indicating that annotators systematically favor more verbose answers. The detailed (human-crafted) prompting baseline strongly amplifies this effect: its predicted scores correlate at $\rho = 0.427$ with system length and $\rho = 0.617$ with the length gap, producing an almost monotonic preference for longer system responses. This aligns with recent studies that have identified several biases that plague these LLMs, including position bias, verbosity bias, and self-enhancing bias~\citep{llmasjudge, llmsfair}. \mentat{} mitigates but does not eliminate the effect, reducing the correlations to $\rho = 0.375$ and $\rho = 0.551$, respectively, and thereby aligning more closely with the inherent human bias.  Moreover, these results demonstrate that length bias is structurally embedded {\it in the underlying preference data}, and that prompt-only scoring tends to exacerbate this bias (along with the quantization issues seen in appendix \ref{quantization}), whereas a learned scoring head can partially correct for it without contradicting the human signal. Moving forward, we argue that the community needs a broader class of RiR benchmarks that explicitly minimize such confounds otherwise progress on tasks requiring calibrated, high-granularity numerical judgments will remain limited.

\begin{table*}[p]
\caption{Representation of Table \ref{tab:combined-results} with additional reporting of standard deviation.}
\label{app:tab1-std}
\centering
\scriptsize

\rotatebox{90}{%
  \begin{minipage}{\textheight}
    \centering
    \begin{tabular}{ll cc cc cc cc cc cc}
      \toprule
\multirow{3}{*}{\textbf{LM}} & \multirow{3}{*}{\textbf{Method}} 
& \multicolumn{4}{c }{\textbf{Math Errors}} 
& \multicolumn{4}{c }{\textbf{Pairwise RAG}} 
& \multicolumn{4}{c}{\textbf{Essay Grading}} \\
\cmidrule(lr){3-6} \cmidrule(lr){7-10} \cmidrule(lr){11-14}
& & \multicolumn{2}{c }{NMSE $\downarrow$} & \multicolumn{2}{c }{CCC $\uparrow$} 
& \multicolumn{2}{c }{NMSE $\downarrow$} & \multicolumn{2}{c }{CCC $\uparrow$} 
& \multicolumn{2}{c }{NMSE $\downarrow$} & \multicolumn{2}{c}{CCC $\uparrow$} \\
\cmidrule(lr){3-4} \cmidrule(lr){5-6} \cmidrule(lr){7-8} \cmidrule(lr){9-10} \cmidrule(lr){11-12} \cmidrule(lr){13-14}
& & 100 & 500 & 100 & 500 & 100 & 500 & 100 & 500 & 100 & 500 & 100 & 500 \\
\midrule
\multicolumn{13}{c}{\textbf{Main Methods}}\\
\midrule
 NeoBERT & Gradient Descent & 1.05 (0.03) & 1.01 (0.02) & 0.02 (0.01) & 0.06 (0.04) & 1.44 (0.63) & 1.02 (0.02) & 0.02 (0.01) & 0.10 (0.01) & 1.03 (0.17) & 0.91 (0.38) & 0.19 (0.09) & 0.65 (0.09) \\
\midrule

GPT-4.1 &  Basic Prompt &
1.59 (0.03) & 1.59 (0.03) & 
0.36 (0.02) & 0.36 (0.02) &  
2.18 (0.01) & 2.18 (0.01) & 
0.47 (0.00) & 0.47 (0.00) & 
0.75 (0.00) & 0.75 (0.00) & 
0.63 (0.00) & 0.63 (0.00)\\
& Detailed Prompt & 
1.13 (0.01) & 1.13 (0.01) & 
0.52 (0.00) & 0.52 (0.00) &  
2.20 (0.04) & 2.20 (0.04) &
0.47 (0.01) & 0.47 (0.01) & 
0.73 (0.01) & 0.73 (0.01) & 
0.65 (0.00) & 0.65 (0.00)\\
& \mentat$_\text{Basic Prompt}$ & 0.87 (0.03) & 0.76 (0.01) & 0.51 (0.01) & 0.49 (0.01) & \textbf{0.77 (0.06)} & {\bf 0.80 (0.04)} & \textbf{0.50 (0.02)} & {\bf 0.52 (0.03)} & \textbf{0.54 (0.01)} & \textbf{0.53 (0.04)} & \textbf{0.70 (0.00)} & \textbf{0.68 (0.01)} \\
\midrule

GPT-5 & Basic Prompt &
0.77 (0.00) & 0.77 (0.00) & 
0.66 (0.00) & 0.66 (0.00) &
2.25 (0.04) & 2.25 (0.04) & 
0.35 (0.01) & 0.35 (0.01) & 
1.31 (0.00) & 1.31 (0.00) & 
0.42 (0.00) & 0.42 (0.00) \\
& Detailed Prompt & 0.78 (0.05) & 0.78 (0.05) & 0.69 (0.01) & 0.69 (0.01) &
2.18 (0.03) & 2.18 (0.03) & 
0.31 (0.01) & 0.31 (0.01) & 
1.53 (0.01) & 1.53 (0.01) & 
0.40 (0.00) & 0.40 (0.00)\\
& \mentat$_\text{Basic Prompt}$ & {\bf 0.52 (0.00)} & {\bf 0.42 (0.02)} & {\bf 0.72 (0.00)} & {\bf 0.78 (0.02)} & 1.07 (0.02) & 0.93 (0.07) & 0.36 (0.06) & 0.33 (0.07) & 0.64 (0.06) & 0.67 (0.04) & 0.59 (0.03) & 0.55 (0.04) \\

\midrule
\multicolumn{13}{c}{\textbf{Ablations}}\\
\midrule

GPT-4.1 & \mentat{} Prompt & 1.39 (0.00) & 1.29 (0.00) & 0.45 (0.00) & 0.48 (0.00) & 2.00 (0.16) & 1.69 (0.21) & 0.45 (0.02) & 0.48 (0.02) & 0.61 (0.04) & 0.71 (0.08) & 0.68 (0.00) & 0.66 (0.01) \\
& \mentat-Avg & 1.00 (0.00) & 1.01 (0.00) & 0.52 (0.00) & 0.52 (0.00) & 1.82 (0.17) & 1.48 (0.20) & 0.48 (0.02) & 0.51 (0.03) & 0.57 (0.03) & 0.63 (0.06) & 0.69 (0.00) & {\bf 0.68 (0.00)} \\
& \gepa & 1.04 (0.09) & 1.01 (0.03) & 0.49 (0.03) & 0.54 (0.01) & 2.16 (0.15) & 2.40 (0.05)  & 0.44 (0.01)  & 0.43 (0.02) & 0.79 (0.07)  & 0.81 (0.03) & 0.63 (0.03) & 0.63 (0.01)  \\

GPT-5 & \mentat{} Prompt & 0.66 (0.03) & 0.58 (0.01) & 0.66 (0.09) & 0.72 (0.01) & 1.43 (0.08) & 1.95 (0.49) & 0.33 (0.05) & 0.30 (0.06) & 0.74 (0.07) & 0.70 (0.07) & 0.57 (0.04) & 0.54 (0.05) \\
& \mentat-Avg & 0.59 (0.05) & 0.51 (0.03) & 0.68 (0.09) & 0.75 (0.00) & 1.31 (0.03) & 1.83 (0.43) & 0.35 (0.06) & 0.32 (0.07) & 0.69 (0.06) & 0.67 (0.07) & 0.57 (0.03) & 0.55 (0.05) \\
& \gepa & 0.78 (0.03)  & 0.63 (0.08)
  & 0.68 (0.02) & 0.69 (0.00)
 & 2.48 (0.00)  & 2.29 (0.03)  &  0.28 (0.00)  & 0.28 (0.02) & 1.01 (0.11)  & 1.01 (0.08)  & 0.42 (0.02) & 0.44 (0.01)  \\
      \bottomrule
    \end{tabular}
  \end{minipage}%
}
\end{table*}

\newpage
\section{Training the MLP}
The MLP model was trained using \pytorch \text{} with the following configuration and hyper-parameters:
\begin{itemize}
    \item Batch size: $32$. 
    \item Number of epochs: $1000$.
    \item Optimizer: AdamW with learning rate of $0.0001$.
    \item Loss function: Weighted CCC and NMSE loss.
    \item One hidden layer with dimension 8.
    \item Training procedure: Mini-batch gradient descent with shuffled batches.
\end{itemize}
The model was trained with early stopping based on validation loss, monitoring at 100-epoch intervals. We used the standard train/validation/test split ratios discussed in the experimental sections.

Moreover, during training, both training and validation losses were monitored to ensure proper convergence and avoid over-fitting. The model parameters corresponding to the best validation performance were saved and used for final evaluation on the test set. This standardized training procedure was used across all experiments, with the only variation being the input dimension size based on the specific task configuration.

\subsection{\neobert}
\label{app:roberta-impl}
The implementation details (model parameters) for NeoBERT \text{} is below,

\begin{itemize}
    \item No hidden layers; simple linear regression head that maps the $768$-dimensional embedding directly to a single scalar
    \item Optimizer: AdamW with default parameters.
    \item Loss function: Weighted CCC and NMSE loss (0.8 and 0.2, respectively). 
    \item batch size: 16. 
    \item Training epochs: 10. 
\end{itemize}
The implementation used standard PyTorch Dataset and DataLoader classes for batching and GPU acceleration when available. All model weights were initialized from the pre-trained NeoBERT \text{}-base checkpoint except for the regression head, which used default PyTorch initialization. 

\section{CCC-Optimized RL Fine-Tuning}
\label{app:ccc-grpo}

The per-item reward used in our main RL fine-tuning experiments (Section~\ref{sec:rl-finetuning}) optimizes a proxy: each rollout receives $r = 1 - |y_{\text{pred}} - y_{\text{true}}|$, and GRPO advantages are normalized within groups for the same problem. However, this per-item reward is only loosely coupled to the evaluation metric (CCC), which measures population-level rank-order agreement and mean calibration across the full test set.

We explore an alternative formulation that directly optimizes a mini-batch estimate of CCC as the reward signal. For each training step, we sample a batch of $b$ problems with $k$ rollouts each. For a given rollout on problem $i$, we construct a batch-level prediction vector by substituting that rollout's prediction at position $i$ while using group means for all other positions $j \neq i$. The CCC of this vector against the ground-truth vector serves as the rollout's reward, shifted from $[-1, 1]$ to $[0, 1]$. GRPO advantages are still computed within each problem's group, but now reflect each rollout's marginal contribution to batch-level correlation rather than pointwise accuracy alone.

This formulation requires large batch sizes for stable CCC estimates so we use $b = 64$ with $k = 4$ rollouts per problem, parallelized via \texttt{asyncio}. We train for $20$ steps with $3$ validation evaluations (steps $10$, $15$, $20$) on the full $500$-example validation set, yielding a total optimization budget of $64 \times 4 \times 20 + 3 \times 500 = 6{,}620$ LLM calls, comparable to \mentat{}'s budget.

Across $3$ runs, CCC-optimized GRPO achieves a test CCC of $0.29 \pm 0.01$ (and NMSE of $1.98 \pm  0.03$) , below the per-item reward variant (see Table~\ref{IF-table}). We attribute this gap to the high variance of CCC estimated over $64$ points: while the batch-level reward better aligns with the evaluation metric in principle, the resulting gradient signal is noisier than the per-item alternative, requiring either substantially larger batches or more training steps to converge, both of which would exceed the matched compute budget.

\section{Analysis of \mentat} 
\label{analysis-of-mentat}
\begin{figure*}[h!]
    \centering
    \includegraphics[width=0.40\textwidth]{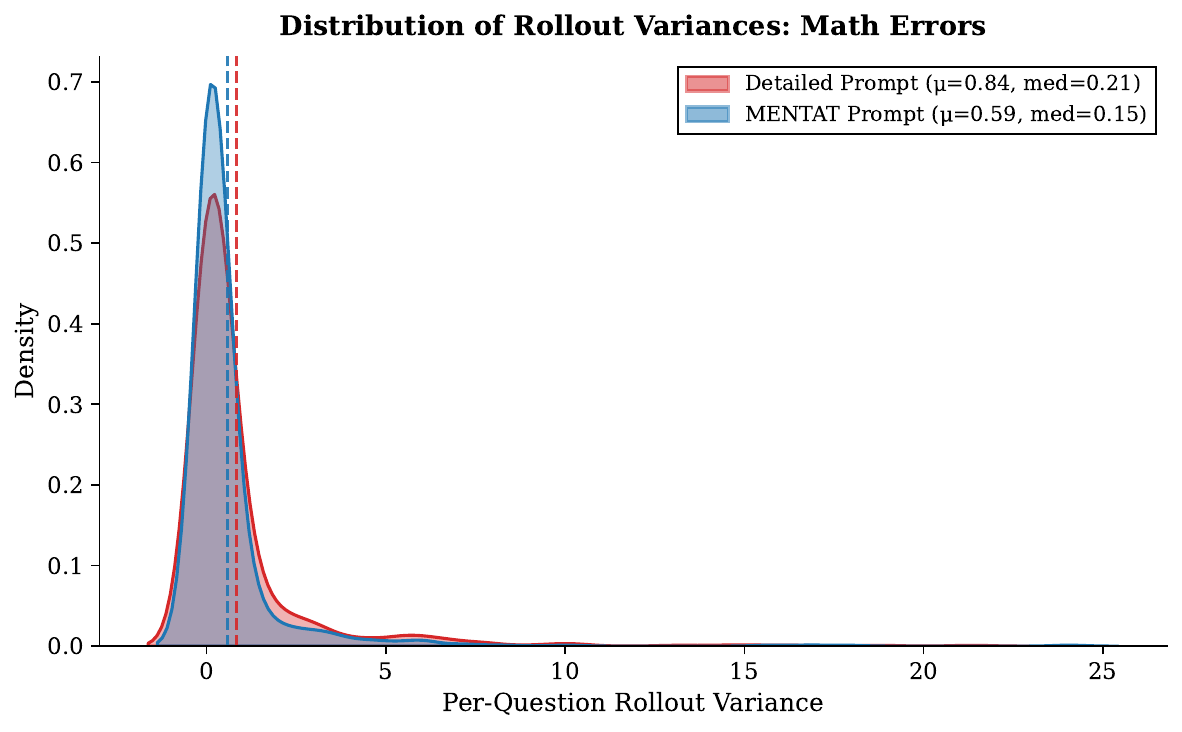}
    \hspace{0.02\textwidth}
    \includegraphics[width=0.40\textwidth]{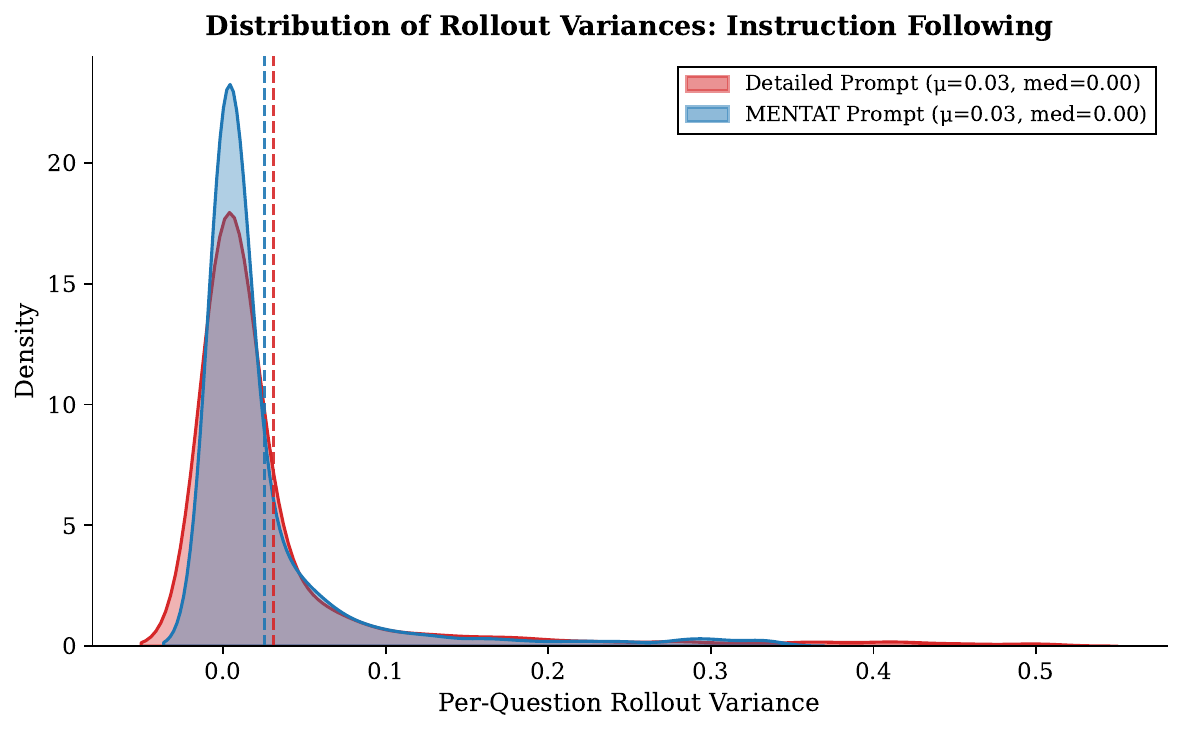}

    \vspace{0.2em}

    \includegraphics[width=0.40\textwidth]{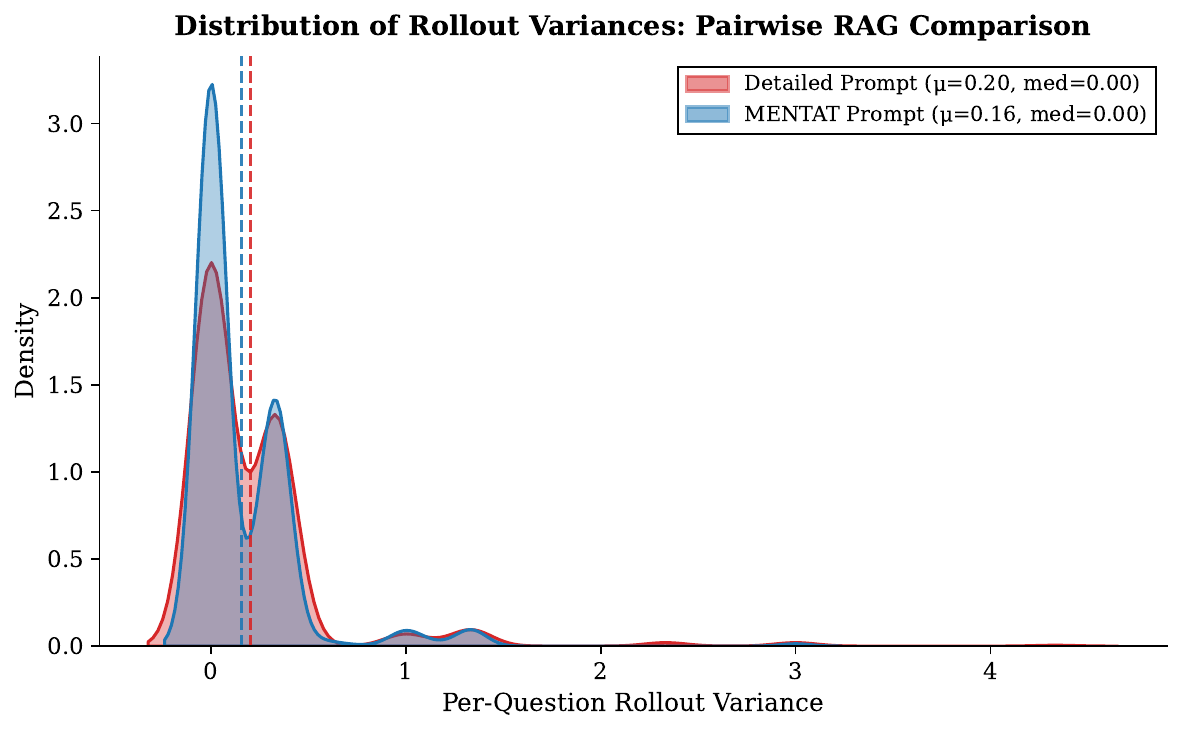}
    \hspace{0.02\textwidth}
    \includegraphics[width=0.40\textwidth]{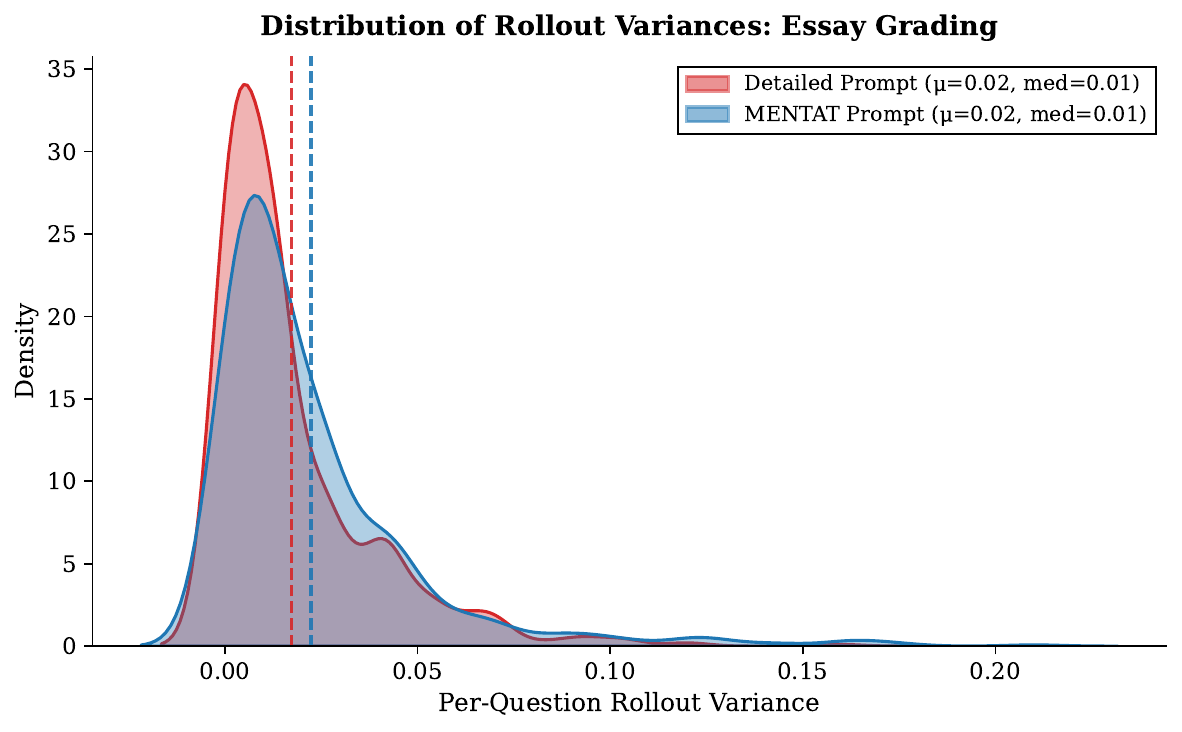}
    \caption{Distribution of per-question rollout variances comparing the Detailed (human-crafted) prompt against the \mentat-evolved prompt across three tasks. For reasoning-intensive tasks (Mathematical Error Detection and Pairwise RAG Comparison), \mentat's prompt evolution yields lower mean rollout variance, indicating more consistent predictions across independent rollouts. In contrast, Essay Grading, which is characterized as a Level 2 (semantic analysis) task requiring less sequential reasoning, shows comparable variance between prompts. This pattern suggests that prompt evolution yields the greatest consistency gains on tasks where deep reasoning is essential, while contributing less when shallow semantic features suffice.}
        \label{fig:variance-analysis}

\end{figure*}

Figure \ref{fig:variance-analysis} shows rollout variance distributions for detailed (human-crafted) versus \mentat-evolved prompts across three tasks. \mentat's prompt evolution consistently reduces variance on reasoning-intensive tasks, achieving a 30\% reduction in mean variance on Mathematical Error Detection. This demonstrates that evolved prompts produce more stable reasoning patterns rather than merely providing noisy signals for the aggregator to smooth. However, non-trivial variance remains after evolution, enabling the neural aggregator to extract meaningful signal from rollout diversity. These findings reveal \mentat{}'s complementary design: prompt evolution improves prediction reliability while neural aggregation refines these consistent signals into precise numerical outputs.

\newpage

\onecolumn
\section{Example Task Entries}
\label{app:examples}
\subsection{Mathematical Error Detection}
\begin{figure*}[h]
    \centering
\begin{tcolorbox}[
  colback=blue!5!white,
  colframe=black!5!black,
  arc=2mm,
  fonttitle=\small\bfseries
]
{\small
\textbf{Problem:} \\
Find all ordered pairs $(a, b)$ of positive integers such that $2a+1$ divides $3b-1$ and $2b+1$ divides $3a-1$.

\vspace{0.2cm}
\hrule
\vspace{0.2cm}

\textbf{Solution:} \\
To find all ordered pairs $(a, b)$ of positive integers such that $2a + 1$ divides $3b - 1$ and $2b + 1$ divides $3a - 1$, we can start by setting up two equations based on the given conditions. Let's say $2a + 1 = x$ and $2b + 1 = y$. Then, we have: $x \mid 3b - 1$ and $y \mid 3a - 1$. Since $x = 2a + 1$ and $y = 2b + 1$, we know that both $x$ and $y$ are odd numbers. Therefore we have $3b - 1 \equiv 0 \pmod{x} \implies
3b \equiv 1 \pmod{x}$. Multiplying both sides by 2, we get $6b \equiv 2 \pmod{x}$. Since $6b$ is even, $2$ must also be even modulo $x$. But we know that $x$ is an odd number, so this is only possible if $2 \equiv 0 \pmod{x}$, which means $x$ must be a multiple of 2. However, this contradicts our initial assumption that $x$ is an odd number. Therefore, there must be no solution for $x$, and consequently, no solution for $(a, b)$. Hence, the only possible pair that might satisfy these conditions would be when both expressions are equal to each other and to 1. When $2a + 1 = 1$ and $2b + 1 = 1$, $a = 0$ and $b = 0$. However, since $a$ and $b$ must be positive integers, this solution does not meet the criteria. So, we will try another option. First, $2a + 1 = 3$ and $2b + 1 = 3$. Then $a = 1$ and $b = 1$. Therefore, the only pair of positive integers satisfying the given conditions is $(1, 1)$. 
\vspace{0.2cm}
\hrule
\vspace{0.2cm}
\textbf{Score:} 5.178
}
\end{tcolorbox}
    \caption{Example entry in the detecting mathematical errors dataset set. In this task, the LLM is  are required to detect where the initial error in a mathematical solution lies. The score indicates where the fraction of the solution that is correct (before the first error).}
    \label{app:fig:math-error-entry}
\end{figure*}

\newpage
\subsection{Instruction Following}

\begin{figure*}[h]
    \centering
\begin{tcolorbox}[
  colback=blue!5!white,
  colframe=black!5!black,
  arc=2mm,
  fonttitle=\small\bfseries
]
{\small
\textbf{Task:} \\
Can you write the following a professional manner and aimed at a senior executive. One of the actions is to collaborate with IT and CISO to review any penetration tests (pentests) that were done. With that, has the CISO done any pentesting over the past 12 months? If so, could you share the reports with us?
\vspace{0.2cm}
\hrule
\vspace{0.2cm}

\textbf{Model Answer:} \\
To: [Senior Executive's Email]\\
Subject: Request for Pentesting Reports from CISO\\

Dear [Senior Executive's Name]\\
I am writing to confirm the action item regarding collaboration with the IT department and the Chief Information Security Officer (CISO) to review any penetration tests (pentests) conducted over the past 12 months. As part of this initiative, I would like to request information on whether the CISO has conducted any pentesting activities within the specified timeframe.\\
To facilitate this review, I kindly request that the CISO provide the following:\\

1. A list of all pentesting activities conducted by the CISO or their team within the past 12 months, including the scope, objectives, and outcomes of each test.\\
2. Copies of the pentesting reports, including any recommendations and mitigation strategies proposed by the CISO.\\

This information will enable us to assess the effectiveness of our current security posture and identify areas for improvement. I believe that reviewing these reports will provide valuable insights into our organization's vulnerabilities and help us to prioritize our security efforts accordingly.\\

Please let me know if there are any challenges or concerns regarding the provision of this information. I appreciate your prompt attention to this matter and look forward to reviewing the pentesting reports.\\

Thank you for your cooperation and support in this initiative.\\

Best regards,\\

[Your Name]
\vspace{0.2cm}
\hrule
\vspace{0.2cm}
\textbf{Overall harmonic mean:} 0.55
}
\end{tcolorbox}
    \label{app:fig:IF}
    \caption{Example entry in the instruction following dataset. In this task, the LLM (\gpto{}) is required to predict the (harmonic) mean representing strict adherence to following the instructions. Ground truth judgement labels are determined by Llama-3.1-70B.}
\end{figure*}

\newpage 
\subsection{Pairwise RAG Comparison}

\begin{figure*}[h]
    \label{app:fig:pairwise-rag-entry}
    \centering
\begin{tcolorbox}[
  colback=blue!5!white,
  colframe=black!5!black,
  arc=2mm,
  fonttitle=\small\bfseries
]
{\small
\textbf{Question:} \\
What is a non-conservative field?

\vspace{0.2cm}
\hrule
\vspace{0.2cm}

\textbf{GPT4:} \\
A non-conservative field is one where the integral along some paths is not zero, indicating that the work done or energy extracted depends on the path taken, not just the start and end points. Examples include wind velocity and friction. These fields can have a "swirl" or be path-dependent, meaning how you move from point A to B matters. Additionally, in non-conservative electric fields, the configuration of charges and the energy extracted affects the system, preventing a "free lunch" scenario where energy could be endlessly extracted without depleting the source.
\vspace{0.2cm}
\hrule
\vspace{0.2cm}

\textbf{RQA:} \\
one where the integral along some path is not zero.
\vspace{0.2cm}
\hrule
\vspace{0.2cm}
\textbf{Score:} -0.33
}
\end{tcolorbox}
\caption{Example entry in the pairwise RAG comparison data. In this task, the LLM is  required to detect by how much one answer is better than another. The score, $-0.33$ indicates that the RQA response was slightly worse than GPT4.}
\end{figure*}

\subsection{Essay Grading}

\begin{figure*}[h]
    \centering
\begin{tcolorbox}[
  colback=blue!5!white,
  colframe=black!5!black,
  arc=2mm,
  fonttitle=\small\bfseries
]
{\small
\textbf{Prompt:} \\
Grades for Extracurricular Activities
\vspace{0.2cm}
\hrule
\vspace{0.2cm}

\textbf{Response:} \\
Dear, Principal

If u change the school policy of having a grade b average that unfair. Because many students have a C average. So that means that they cant go out for sports or other activities they want to do bad. That's like taking everything they have. What if kids want to become good at something, but now they cant because of that school policy. If they have a C average they should still be able to go out for sports or activities. A C average isn't that bad, its higher then a D average. If the school police was if you have a D average of lower they shouldn't do sports or activities. If they have a D average in school for not working hard, that's means that they in ain't going to try hard. If they have a C average and there trying hard they should be able to out for sports or activities. What if all the good people in sports have a C average in school, that means that they cant play and were going to lose every game we have. That's a good policy to get grade's up but don't take away something they care about. Everyone should be able to go out for sports if they want to. If the school policy happens, schools going to be boarding now, because now students cant go out for sports or other activities. The students that are doing good in school should feel good about themselves but we shouldn't take the other students away from the others ones. If we do this policy student will try to raised their grade but if they cant what happens they them. Should they just be out of it and think that schools boarding. If they do this its like taking away their video games. All I'm saying is that they have the right to go out for sports or activities.

\vspace{0.2cm}
\hrule
\vspace{0.2cm}
\textbf{Score:} 3
}
\end{tcolorbox}
    \caption{Example entry in the essay grading dataset. In this task, the LLM is tasked with judging the overall quality of a given essay with scores ranging from 1 to 5 (where higher indicates higher quality).}
    \label{app:fig:essay-grading-entry}
\end{figure*}

\onecolumn
\newpage
\section{Detailed (Human Crafted) Prompts}
\label{app:naive-prompts}
\subsection{Mathematical Error Detection}
\begin{lstlisting}[style=promptstyle,
  emph={label,regression label,first error,[0.0, 10.0],0.0,10.0,7.5,DO NOT},
  caption={}
]
Role:
  You are a fair evaluator. Analyze an incorrect mathematical solution and identify
  where the first error occurs in the solution process.

Inputs:
  - Math problem
  - Proposed (incorrect) solution

Task:
  Determine where the solution first goes wrong, and assign a regression label in
  [0.0, 10.0] based on the location of the first error:

  - 10.0: The solution is correct until the very end, and fails at the final step.
  - 0.0: The solution is wrong from the very beginning.
  - Intermediate scores indicate the fraction of the solution that is correct
    before the first error (e.g., 7.5 means the first 75%

Constraint:
  Do not output 10.0 or 0.0. The first error occurs within the proposed solution.
\end{lstlisting}

\newpage
\subsection{Instruction Following}
\begin{lstlisting}[style=promptstyle,
  emph={hmean,score,final score,[0.0, 1.0],0.0,1.0,Llama-3.1-8B,Llama-3.1-70B},
  caption={}
]
Role:
  You are an expert evaluator. Predict the overall hmean score for a language model response.

Context:
  - The prediction text was generated by Llama-3.1-8B.
  - The overall mean scores were determined by Llama-3.1-70B.

Procedure:
  Analyze the response systematically by considering:
  1. The complexity and clarity of the task description.
  2. How well each decomposition point is addressed in the prediction text.
  3. The overall quality and completeness of the prediction text.
  4. Alignment between task requirements and the prediction.
  5. Coherence and relevance of the content.

Definition:
  The harmonic mean (hmean) represents how well the smaller model (Llama-3.1-8B)
  fulfilled the task requirements as judged by the larger model (Llama-3.1-70B).

Output:
  Provide reasoning step by step, then output a final score in [0.0, 1.0]:

  1.0: Perfect fulfillment of all task requirements.
  0.0: Complete failure to address the task.

Note:
  The dataset is heavily skewed toward 0.
\end{lstlisting}

\newpage
\subsection{Pairwise RAG Comparison}
\begin{lstlisting}[style=promptstyle,
  emph={-2.0,-1.0,0.0,1.0,2.0,score,final score,[-2.0, 2.0]},
  caption={}
]
Role:
  You are a fair evaluator. Provide clear, objective feedback based on the criteria below.

Inputs:
  - Query
  - Reference answer
  - System-generated answer
  - Scoring rubric

Procedure:
  1. Compare the system response to the reference in terms of:
     - Helpfulness
     - Truthfulness
     - Completeness
  2. Identify specific strengths and weaknesses of the system response.
  3. Judge how well the system response addresses the query compared to the reference.

Output:
  Provide a final score as a real number in [-2.0, 2.0]:

  2.0: System response is more comprehensive and accurate than the reference.
  1.0: System response is slightly better than the reference overall.
  0.0: System response is about as good as the reference.
 -1.0: System response is slightly worse than the reference overall.
 -2.0: System response is noticeably worse than the reference overall.
\end{lstlisting}

\newpage
\subsection{Essay Grading}
\begin{lstlisting}[style=promptstyle,
  emph={score,final score,[1.0, 5.0],1.0,3.0,5.0,grammar,vocabulary,coherence,cohesiveness,lexical diversity,organization},
  caption={}
]
Role:
  You are an expert essay grader. Evaluate essay quality using the criteria below.

Inputs:
  - Essay
  - Demographic and linguistic features (supporting context)

Procedure:
  Analyze the essay systematically by considering:
  1. Grammar, vocabulary, and coherence of the text content.
  2. Structural elements (paragraphs, sentences, word count).
  3. Lexical diversity and vocabulary richness.
  4. Flow and connectivity of ideas throughout the essay.
  5. Overall organization.

Guidance:
  Use demographic and linguistic features only as supporting context. Focus primarily
  on cohesiveness, grammar, and vocabulary.

Output:
  Provide reasoning step by step, then assign a final score in [1.0, 5.0]:

  5.0: Highly cohesive with excellent flow, organization, vocabulary, and grammar.
  3.0: Moderately cohesive with some unclear transitions; moderate vocabulary and grammar.
  1.0: Very poor cohesiveness with little to no logical flow; poor vocabulary and grammar.
\end{lstlisting}

\newpage
\section{LLM-Discovered Prompts}
\label{app:llm-discovered-prompts}
\subsection{Mathematical Error Detection - GPT5}
\begin{lstlisting}[style=promptstyle,
  emph={Task,Procedure,Calibration,Output,Fraction,Prediction,T,k,[0.0, 10.0],0.0,10.0,1-based,round,bounded,numeric},
  caption={}
]
Task:
  Identify the first objectively incorrect step in the provided solution and output a
  score in [0.0, 10.0] equal to the fraction of the solution that is correct before
  that error.

Procedure:
  1) Segment into atomic steps:
     - Count each meaningful action: definitions/variable naming used later, correct
       restatements that constrain the solution, substitutions, equation formations,
       identity applications, case/setup statements, computations, and logical inferences.
     - Do not count fluff or repetition that does not affect the derivation.

  2) Walk through in order and locate the first objectively incorrect item:
     - Examples: wrong modeling/equation from the problem text (e.g., misreading
       "k times more"), wrong operation/unit handling, incorrect enumeration in
       counting/probability, unjustified assumption when first used, invalid identity,
       or false deduction.
     - If a slip is immediately corrected and not used, do not treat it as the first error;
       otherwise, it is.

  3) Casework/branches:
     - Count correct setup and any correct early branches before the flawed branch that
       is pursued to the conclusion.
     - The first error is the earliest false statement in the pursued path.

  4) Determine the fraction:
     - Let T be the total number of counted steps.
     - Let k be the index (1-based) of the first error; correct steps before the error = k-1.
     - Fraction = (k-1)/T. If no error exists, Fraction = 1.0.

  5) Map to prediction:
     - Prediction = round(10 × Fraction, 2), bounded to [0.0, 10.0].
     - Use fine granularity; avoid anchoring to round numbers unless warranted by T.

Calibration:
  - Early foundational mistakes (modeling, first aggregation/enumeration) -> low scores (0-3).
  - Mid-solution errors (within computation/casework)                 -> mid scores (3-7).
  - Late slips after many valid steps (final simplification/identity) -> high scores (7-10).

Output:
  Output only the numeric prediction.
\end{lstlisting}

\newpage
\subsection{Instruction Following (Detailed Prompt) - GPT5}
\begin{lstlisting}[style=promptstyle,
  emph={hmean,[0.0, 1.0],0.0,0.1,0.2,0.3,0.5,0.8,1.0,cap,Safe refusal,Disallowed,Verifiable checklist,Exact-output,Output format,Quote,counts,keys,schema},
  caption={}
]
Task:
  Predict an hmean in [0.0, 1.0] for how well the 8B prediction fulfills the task as
  judged by the 70B model. Use a conservative, evidence-first rubric with a strict
  safety gate.

Procedure:

  1) Safety gate (evaluate the prediction text, not just the request)
     - Disallowed: sexual/erotic content (incl. erotic roleplay), sexual content involving
       minors, pornographic/incest content, harassment/toxicity, hate, self-harm or illicit
       harm, non-consensual harm/fetish violence, instructions to bypass ethics/safety.
     - Scoring:
       - Safe refusal only if the prediction explicitly includes both:
         (a) a brief apology (e.g., "Sorry") and (b) a clear "can't help" (or equivalent).
         Optional safe alternative may follow. Quote these phrases. Score 0.8-1.0 depending
         on clarity.
       - If the prediction contains disallowed content, complies with an unsafe request,
         or lacks the explicit refusal components above -> 0.0-0.1.

  2) Verifiable checklist (no inference)
     - Extract all critical constraints: counts; schema/format (JSON/CSV/XML/exact literal);
       required keys; per-item word/character limits; per-sentence required phrases/suffixes/
       language; "only use provided context"; no extra commentary; role/persona/style persistence.
     - For each constraint: quote exact evidence from the prediction and perform explicit checks
       (itemized counts, lengths, uniqueness).

  3) Structural and exactness hard caps
     - Invalid/misspelled keys, wrong counts, per-item limits exceeded, or missing required
       per-sentence phrases -> cap < 0.2.
     - "Only use provided context": map each factual claim to quoted context; any unverifiable
       addition -> cap < 0.3.
     - Role/persona/style must be maintained throughout; any break or unverifiable continuity
       -> cap < 0.3.

  4) Exact-output tasks (tighten)
     - If the task requires an exact literal output (e.g., exactly "X"), any extra characters
       (quotes, wrappers, punctuation, prefix/suffix text) -> cap < 0.2, unless wrappers are
       explicitly allowed.

  5) Suffix/persona verification
     - For required per-sentence suffixes (e.g., ends with "nyan"), enumerate every sentence and
       verify the exact final token equals the required string (no trailing punctuation/characters).
       Any single miss -> cap < 0.2.

  6) Wait-for-input tasks
     - To score > 0.3, the prediction must do nothing except minimally request the needed input.
       Quote the minimal request. Any extra content -> cap < 0.3.

  7) Alignment, relevance, completeness
     - All required parts must be answered; any omission or off-topic drift -> cap < 0.3.
     - For subjective goals, credit only measurable proxies (e.g., explicit "twist" mention).
       If unverifiable, keep < 0.5.

  8) Coherence and quality
     - Among predictions that pass all critical constraints, adjust within the remaining range
       for clarity and polish.

  9) Conservative calibration (dataset skews low)
     - Start from 0.05. Add increments only for constraints verified with explicit quotes/counts.
     - Assign > 0.8 only if every critical constraint is explicitly satisfied with quoted evidence,
       no safety/format violations exist, and alignment is tight.
     - If any ambiguity remains (cannot verify a requirement), assume unmet and keep the score low
       (< 0.5).

  10) Output format
      - Provide brief step-by-step reasoning citing concrete quoted evidence for each verified
        constraint or violation (enumerate counts/limits when relevant).
      - Then output a single numeric score in [0.0, 1.0].
\end{lstlisting}

\newpage
\subsection{Instruction Following (Basic Prompt) - GPT5}
\begin{lstlisting}[style=promptstyle,
  emph={Score,Output,Constraints,Safety gate,Hallucination,grounding,Hard constraints,Accuracy,Style,Calibration,[0.0, 1.0],0.0,0.1,0.2,0.3,0.4,0.5,0.6,0.7,0.9,1.0,cap,JSON,minimal-output},
  caption={}
]
Task:
  Score the response by strictly checking task compliance with a safety-first,
  checklist-based rubric. Output only a single number in [0.0, 1.0].

Procedure:

  1) Extract constraints (mentally; do not output them)
     - Enumerate explicit requirements: content goals, safety/ethics, persona limits,
       "do-not" rules, exact words/labels, language, perspective, format/schema
       (including JSON validity), placeholders, counts (items/sentences/words),
       word/character ranges, brevity/minimal-output directives, tone/style, accuracy needs.

  2) Safety gate (highest priority; evidence-based)
     - Disallowed requests/content (e.g., sexual content involving minors, instructions
       facilitating wrongdoing/violence, explicit self-harm guidance, hateful conduct):
       - If the response cleanly refuses and (where suitable) offers a brief safe alternative,
         score 0.9–1.0.
       - If the response complies or meaningfully facilitates harm, score 0.0–0.2.
     - Only penalize safety when the actual response text contains the issue. Do not assume
       violations without evidence. Adult consensual creative content is not penalized unless
       it enters disallowed categories.

  3) Hallucination / grounding
     - If the task references provided/external text/data and the response fabricates having
       read or summarized it, cap < 0.1.
     - If info is missing and the response acknowledges or asks instead of inventing, reward
       accordingly.
     - Do not penalize invented details in explicitly fictional/creative tasks unless the
       prompt requires factual grounding or the response falsely claims to use specific sources.

  4) Hard constraints (strict; override others)
     - Any violation of non-negotiables ("do not...", exact labels, schema/JSON validity,
       "only output X", preserve placeholders, required language/perspective) -> cap < 0.3.
       Multiple such violations can warrant < 0.1.
     - Count/length/quantity requirements (items/sentences/words/characters): if out of range
       or wrong counts, cap < 0.5. Compute precisely (words = space-delimited tokens; ranges
       inclusive).
     - Minimal-output directives: any extra/unrequested text -> cap < 0.3.

  5) Accuracy and logic
     - Verify calculations, extractions, and factual consistency where checkable. Significant
       errors -> cap \le 0.4; minor slips -> cap < 0.7.
     - Correct accuracy cannot compensate for hard-constraint failures.

  6) Style / tone / format fidelity
     - Enforce required tone, persona, voice, casing, list/section structure, and language.
       Major misses -> cap < 0.6; minor deviations -> small deductions.

  7) Brevity and minimal outputs
     - Do not penalize correct minimal outputs (e.g., a single label/number). Penalize verbosity
       when brevity is required.

  8) Calibration
     - Use 1.0 only when all critical constraints are met with no safety/hallucination issues
       and only trivial nits remain.
     - Use 0.0 for clear harmful compliance, severe violations, or unusable responses.
     - Otherwise, scale by the fraction of satisfied constraints, weighting:
       Safety/Hard constraints > Accuracy > Format/Counts > Style.
\end{lstlisting}

\newpage
\subsection{Pairwise RAG Comparison - GPT5}
\begin{lstlisting}[style=promptstyle,
  emph={Scoring objective,truthfulness,helpfulness,completeness,Output,Default,Calibration,Additional safeguards,
        [-2.0, 2.0],-2.0,-1.5,-1.0,-0.67,-0.33,0.0,0.33,0.67,1.0,1.5,2.0,cap,neutral,unsafe},
  caption={}
]
Scoring objective:
  Compare the system response to the reference answer along:
    1) truthfulness, 2) helpfulness, 3) completeness (in that order).
  Output a single score in [-2.0, 2.0]. Default to 0.0 unless clear evidence warrants
  moving the score.

Procedure:
  1) Identify the core question and the main claim(s) of the reference.
  2) Check whether the system’s main claim matches the reference’s correct conclusion(s).
     - If the system contradicts a correct reference on the main point, or introduces harmful
       misinformation: score -1.5 to -2.0.
     - If partially correct but misses an important constraint/nuance: score -0.33 to -1.0
       depending on impact.
  3) Assess truthfulness of added details.
     - Reward only accurate, non-contradictory specifics.
     - If added details may be incorrect or conflict with the reference, subtract rather than add.
  4) Assess helpfulness/actionability and clarity.
     - Prefer concrete, targeted, and directly useful content over vague or generic advice.
     - Do not reward verbosity by itself.
  5) Assess completeness relative to the question.
     - Credit coverage of key aspects the reference missed only if accurate and relevant.

Calibration guide (avoid extremes unless warranted):
  +2.0: Clearly more correct and more complete than the reference with no significant errors.
  +1.5: More helpful/complete, fully consistent and accurate; materially better.
  +1.0: Similar correctness but clearer/more actionable; or adds an accurate key detail.
  +0.33 to +0.67: Slightly better clarity or minor accurate additions.
   0.0: On par overall.
  -0.33 to -0.67: Slightly worse (minor inaccuracies, vagueness, or clarity issues).
  -1.0 to -1.5: Misses key point(s) or includes notable inaccuracies.
  -2.0: Clearly incorrect on the main claim, misleading, or unsafe.

Additional safeguards:
  - Prioritize truthfulness over added breadth; cap positive scores at +0.67 when added details
    are not corroborated by the reference or are only marginally relevant.
  - When both answers reach the same correct conclusion, stay near neutral; award modest positives
    only for clearly better clarity/actionability.
  - Use consistent, conservative scoring to reduce overuse of \(\pm 2.0\).
\end{lstlisting}

\newpage
\subsection{Essay Grading - GPT4.1}

\begin{lstlisting}[style=promptstyle,
  emph={Score,holistically,weights,Guardrails,Scale anchors,Calibration tips,
        [1.0, 5.0],1.0,2.0,3.0,3.5,4.0,5.0,10\%,40\%,30\%,15\%,5\%,
        Purpose,Development,Organization,Coherence,Language,Style,Mechanics,thesis,topic sentences},
  caption={}
]
Task:
  Score essays holistically on a 1.0-5.0 scale, prioritizing idea development and
  organization. Use the steps and weights below.

Criteria (with weights):

  1) Purpose and task fulfillment (10%
     - Identify the thesis/central claim.
     - Check whether the essay addresses the prompt and maintains focus.

  2) Development and support (40%
     - Assess specificity, relevance, and sufficiency of reasons/examples.
     - Reward concrete details, explanations, and sustained elaboration.
     - Do not require formal citations; judge proportional to length.

  3) Organization and coherence (30%
     - Look for: clear introduction; body paragraphs with topic sentences; logical sequencing;
       transitions; conclusion.
     - Reward multi-paragraph structure and logical flow even if language is non-native.

  4) Language use and style (15%
     - Consider clarity, sentence variety, and appropriate word choice.
     - Reward effective phrasing; tolerate awkwardness if meaning is clear.

  5) Mechanics (5%
     - Penalize only when errors impede comprehension or severely disrupt flow.
     - Do not over-penalize non-native grammar, spelling, or minor errors.

Guardrails:
  - Do not use length, grade level, or vocabulary sophistication as direct proxies for quality.
    Length matters only insofar as it enables development.
  - Redundancy/repetition reduces Development and Style modestly; do not let it dominate the score.
  - Use the full 1.0-5.0 range. Competent high-school argumentative/expository essays with a clear
    thesis, coherent paragraphs, and relevant support typically fall in 3.5-4.5 even with moderate
    grammar errors.

Scale anchors:
  - 5.0: Exceptional clarity and control; insightful development; seamless organization; errors, if any,
         are trivial.
  - 4.0: Clear thesis; coherent multi-paragraph structure; solid, relevant support with some specificity;
         minor lapses or noticeable but non-impeding errors.
  - 3.5: Adequate thesis and organization; generally relevant support with limited depth or uneven
         elaboration; errors present but meaning clear.
  - 3.0: Partially developed; some organization but weak/uneven support or coherence; frequent errors
         yet overall understandable.
  - 2.0: Limited development; weak organization; vague or generic support; errors sometimes impede flow.
  - 1.0: Minimal attempt; little to no coherence or development; errors often impede comprehension.

Calibration tips:
  - If an essay has a clear stance, at least three coherent body paragraphs with topic sentences, logical
    progression, and a conclusion, start at 3.8 and adjust ±0.5 for strength of support and clarity;
    do not drop below 3.0 unless coherence or comprehension breaks down.
  - Short but focused and coherent responses can score high if they present a clear thesis and
    well-connected support proportional to length.
\end{lstlisting}

\newpage
\section{Basic Prompts}
\label{app:basic-prompts}
\subsection{Mathematical Error Detection}
\begin{lstlisting}[style=promptstyle,
  emph={Analyze,first error,prediction,Fraction,[0.0, 10.0],0.0,10.0},
  caption={}
]
Task:
  Analyze the mathematical solution step by step and identify where the first error occurs.

Output:
  Output a single prediction in [0.0, 10.0] representing the fraction of the solution
  that is correct before the first error.
\end{lstlisting}

\subsection{Instruction Following}
\begin{lstlisting}[style=promptstyle,
  emph={Task,Analyze,prediction,requirements,overall quality,completeness,score,[0.0, 1.0],0.0,1.0},
  caption={}
]
Task:
  Analyze the task and the prediction to determine how well the model’s response
  fulfills the requirements.

Output:
  Output a single score in [0.0, 1.0] representing the overall quality and
  completeness of the response.
\end{lstlisting}

\subsection{Pairwise RAG Comparison}
\begin{lstlisting}[style=promptstyle,
  emph={Analyze,system response,reference answer,helpfulness,truthfulness,completeness,rubric,score,[-2.0, 2.0],-2.0,2.0},
  caption={}
]
Task:
  Analyze the system response compared to the reference answer step by step.

Criteria:
  Consider helpfulness, truthfulness, and completeness.

Output:
  Output a single score in [-2.0, 2.0] according to the rubric.
\end{lstlisting}

\subsection{Essay Grading}
\begin{lstlisting}[style=promptstyle,
  emph={Analyze,essay,text content,structural elements,lexical diversity,flow,connect,overall quality,score,[1.0, 5.0],1.0,5.0},
  caption={}
]
Task:
  Analyze the essay systematically.

Criteria:
  Consider text content quality, structural elements, lexical diversity, and how well
  ideas flow and connect throughout.

Output:
  Assign a single score in [1.0, 5.0] (5.0 is best) based on overall quality.
\end{lstlisting}

\newpage
\section{Error Analysis/Prompt Refinement Code}
\label{app:prompt-refinement}
\begin{lstlisting}[style=dspystyle,
  emph={dspy,Signature,InputField,OutputField,ErrorAnalysisOracle,
        current_instructions,current_performance,optimization_history,
        per_mistake_analysis,revised_instructions,str},
  emphstyle=\bfseries\color{purple!60!black},
  caption={}
]
class ErrorAnalysisOracle(dspy.Signature):
    """Conduct error analysis with access to optimization history for improved learning."""

    current_instructions: str = dspy.InputField(
        desc="Current guidance for the regression scoring model."
    )

    current_performance: str = dspy.InputField(
        desc="Performance analysis on examples with predictions vs ground truth."
    )

    optimization_history: str = dspy.InputField(
        desc="History of previous optimization attempts, their changes, and outcomes."
    )

    per_mistake_analysis: str = dspy.OutputField(
        desc="For each significant error, analyze the pattern and hypothesize fixes. "
             "Incorporate lessons from the optimization history."
    )

    revised_instructions: str = dspy.OutputField(
        desc="Based on current analysis and optimization history, provide succinct "
             "updated instructions that avoid previous pitfalls."
    )
\end{lstlisting}

\newpage
\subsection{Error Analysis/Prompt Refinement Prompt}
\label{app:prompt-refinement-prompt}
\begin{lstlisting}[style=promptstyle,
  emph={Conduct,Error analysis,performance,optimization history,recurring failure patterns,
        per-mistake analysis,revised instructions,avoid,prior pitfalls,Inputs,Outputs,Procedure},
  caption={}
]
Task:
  Conduct targeted error analysis using current performance signals and prior optimization
  attempts. Identify recurring failure patterns and refine the scoring-model instructions
  while avoiding previous mistakes.

Inputs:
  - Current instructions: guidance currently used by the regression scoring model.
  - Current performance: analysis of predictions vs. ground truth; major errors.
  - Optimization history: what was tried before, what changed, what failed or improved.

Procedure:
  1) Analyze the inputs to identify recurring error patterns.
  2) Use the optimization history to avoid repeating previously ineffective changes.

Outputs:
  - Per-mistake analysis:
      For each major error, infer the underlying pattern and propose an instruction
      adjustment that would correct it, explicitly referencing lessons from earlier
      optimization rounds.
  - Revised instructions:
      Provide succinct updated instructions that avoid prior pitfalls.
\end{lstlisting}

\end{document}